\documentclass[Afour,sageh,times]{sagej}

\usepackage{moreverb,url}

\usepackage[colorlinks,bookmarksopen,bookmarksnumbered,citecolor=red,urlcolor=red]{hyperref}

\usepackage{amsmath}
\usepackage{amssymb}

\usepackage{tabularx}
\usepackage{graphicx}
\usepackage{bm}
\usepackage{color}
\usepackage{float}
\usepackage{units}
\usepackage{mathtools}
\usepackage{url}
\usepackage{hyperref}
\usepackage{balance}
\usepackage{amsmath}
\usepackage{lipsum}
\usepackage{soul}

\makeatletter 
\def\endfigure{\end@float} 
\def\endtable{\end@float}
\makeatother

\usepackage{subcaption}
\usepackage{caption}

\newcommand{\jma}[1]{\textcolor{black}{#1}}

\newcommand{\rev}[1]{\textcolor{black}{#1}}

\newcommand\BibTeX{{\rmfamily B\kern-.05em \textsc{i\kern-.025em b}\kern-.08em
T\kern-.1667em\lower.7ex\hbox{E}\kern-.125emX}}

\begin{document}

\runninghead{Li et al.}

\title{Dynamic Loco-manipulation on HECTOR: 
Humanoid for Enhanced ConTrol and Open-source Research}

\author{Junheng Li\affilnum{1}, Junchao Ma\affilnum{1}, Omar Kolt\affilnum{1}, Manas Shah\affilnum{1}, and Quan Nguyen\affilnum{1}}

\affiliation{\affilnum{1} Department of Aerospace and Mechanical Engineering, University of Southern California, USA}

\corrauth{Junheng Li, Department of Aerospace and Mechanical Engineering, University of Southern California, CA 90089, USA.}

\email{junhengl@usc.edu}

\begin{abstract}
Despite their remarkable advancement in locomotion and manipulation, humanoid robots remain challenged by a lack of synchronized loco-manipulation control, hindering their full dynamic potential.
In this work, we introduce a versatile and effective approach to controlling and generalizing dynamic locomotion and loco-manipulation on humanoid robots via a Force-and-moment-based Model Predictive Control (MPC). Specifically, we proposed a simplified rigid body dynamics (SRBD) model to take into account both humanoid and object dynamics for humanoid loco-manipulation. This linear dynamics model allows us to directly solve for ground reaction forces and moments via an MPC problem to achieve highly dynamic real-time control. 
Our proposed framework is highly versatile and generalizable. We introduce HECTOR (Humanoid for Enhanced ConTrol and Open-source Research) platform to demonstrate its effectiveness in hardware experiments.
With the proposed framework, HECTOR can maintain exceptional balance during double-leg stance mode, even when subjected to external force disturbances to the body or foot location. In addition, it can execute 3-D dynamic walking on a variety of uneven terrains, including wet grassy surfaces, slopes, randomly placed wood slats, and stacked wood slats up to $6 \:\unit{\bf{cm}}$ high with the speed of $0.6 \:\unit{\bf{m/s}}$. In addition, we have demonstrated dynamic humanoid loco-manipulation over uneven terrain, carrying $2.5\: \unit{\bf{kg}}$ load. HECTOR simulations, along with the proposed control framework, are made available as an open-source project. (\url{https://github.com/DRCL-USC/Hector_Simulation}). 
\end{abstract}


\maketitle




\section{Introduction}
\label{sec:Introduction}

\subsection{Motivations}

\begin{figure*}[!t]
\vspace{0.2cm}
     \centering
     \:\begin{subfigure}[b]{0.415\textwidth}
         \centering
         \includegraphics[clip, trim=0cm 1.25cm 2.4cm 1cm, width=1\columnwidth]{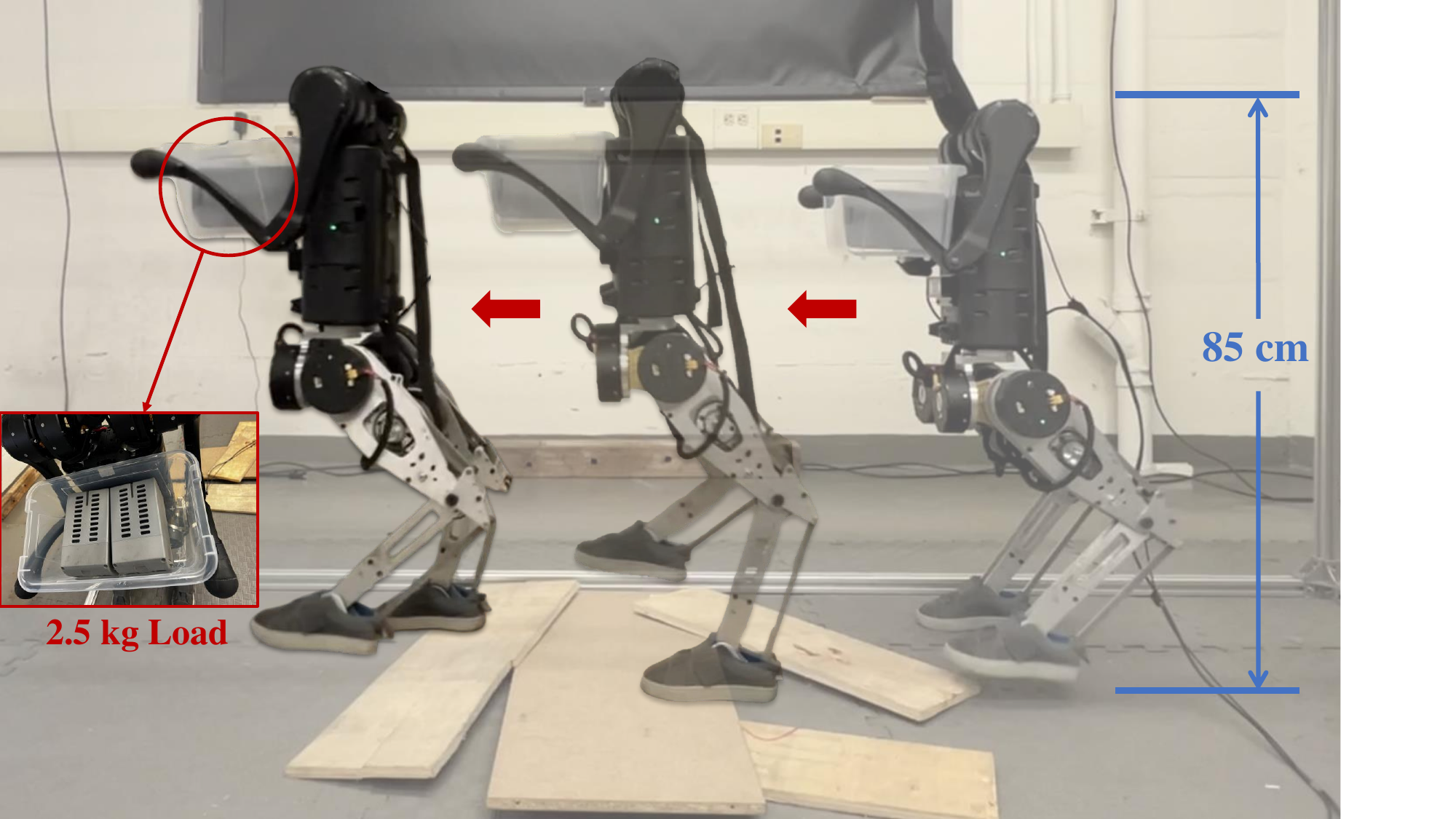}
         \caption{\centering}
         \label{fig:loco_hardware}
     \end{subfigure} 
     \begin{subfigure}[b]{0.16\textwidth}
         \centering
         \includegraphics[clip, trim=1cm 0cm 11cm 0cm, width=1\columnwidth]{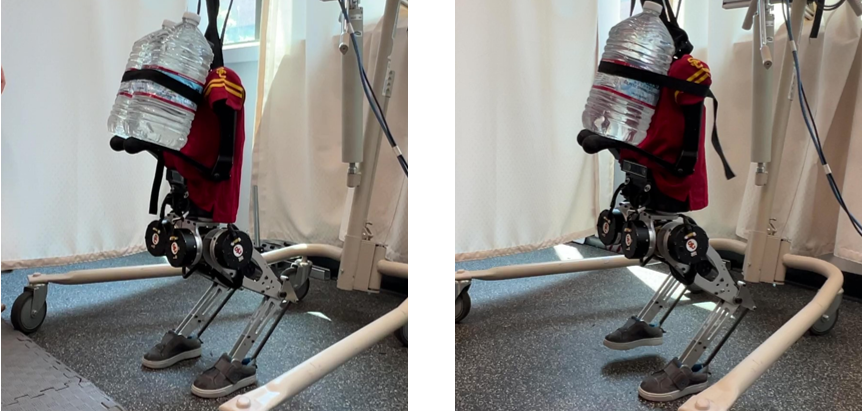}
         \caption{\centering}
         \label{fig:Carrying}
     \end{subfigure} 
     \:\begin{subfigure}[b]{0.16\textwidth}
         \centering
         \includegraphics[clip, trim=11cm 0cm 1cm 0cm, width=1\columnwidth]{NewFigures/weight_carrying.PNG}
         \caption{\centering}
         \label{fig:Carrying2}
     \end{subfigure} 
     \:\begin{subfigure}[b]{0.16\textwidth}
         \centering
         \includegraphics[clip, trim=14.7cm 0cm 2cm 0cm, width=1\columnwidth]{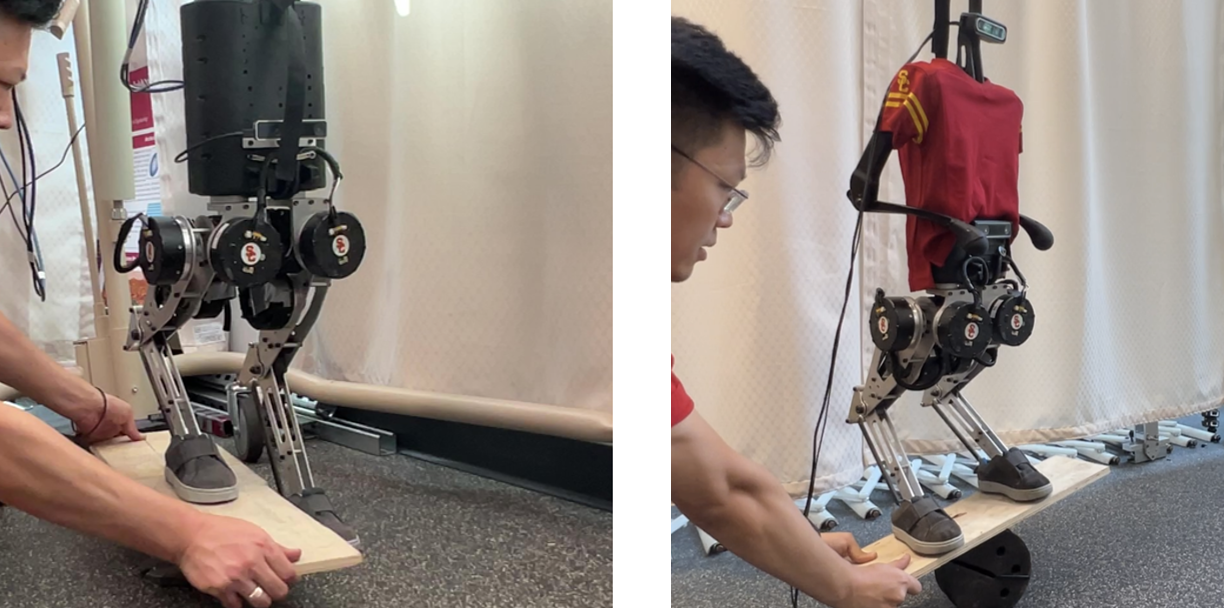}
         \caption{\centering}
         \label{fig:balancing}
     \end{subfigure} 
     \vspace{0.1cm}
     \\
     \quad 
     \begin{subfigure}[b]{0.20\textwidth}
         \centering
         \includegraphics[clip, trim=0cm 0cm 17.5cm 0cm, width=1\columnwidth]{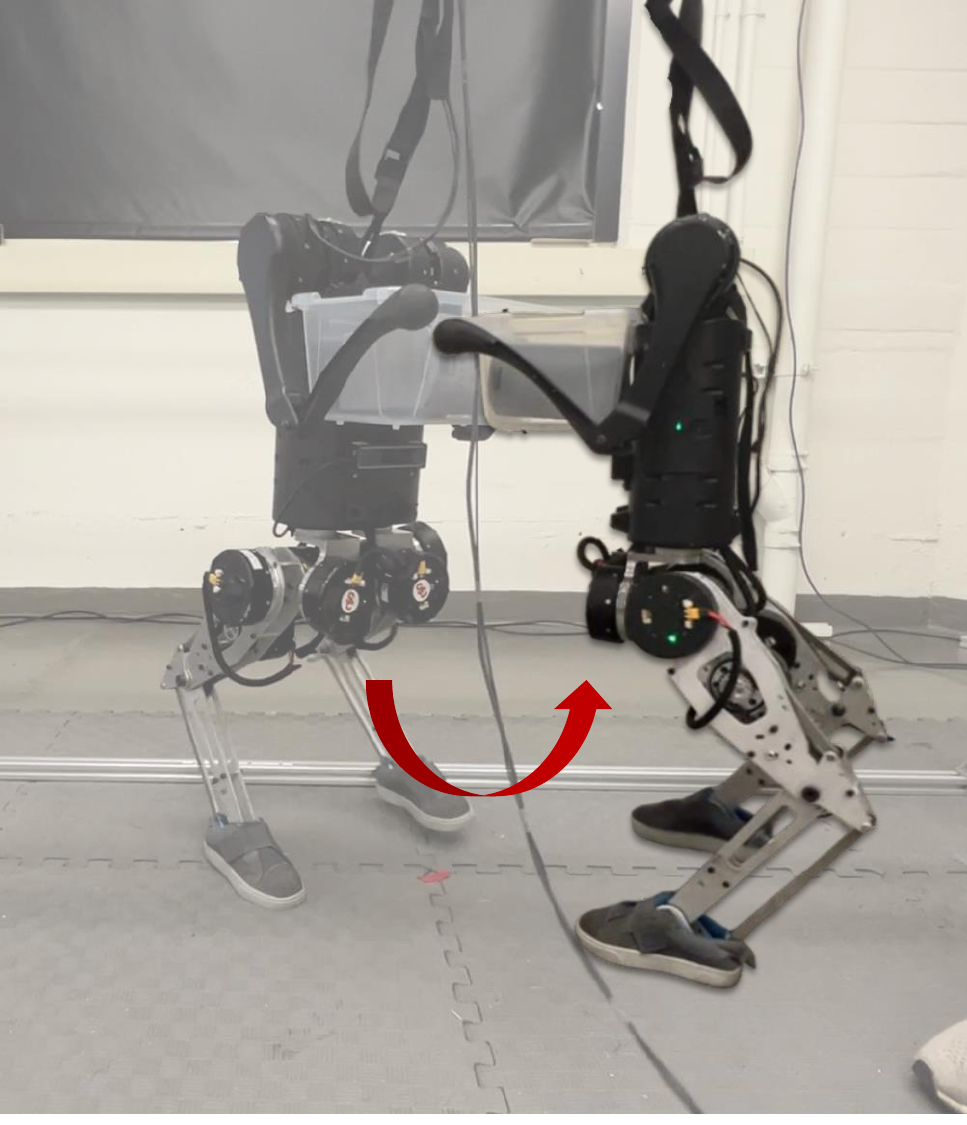}
         \caption{\centering}
         \label{fig:turning}
     \end{subfigure} 
     \begin{subfigure}[b]{0.315\textwidth}
         \centering
         \includegraphics[clip, trim=0cm 0cm 8cm 0cm, width=1\columnwidth]{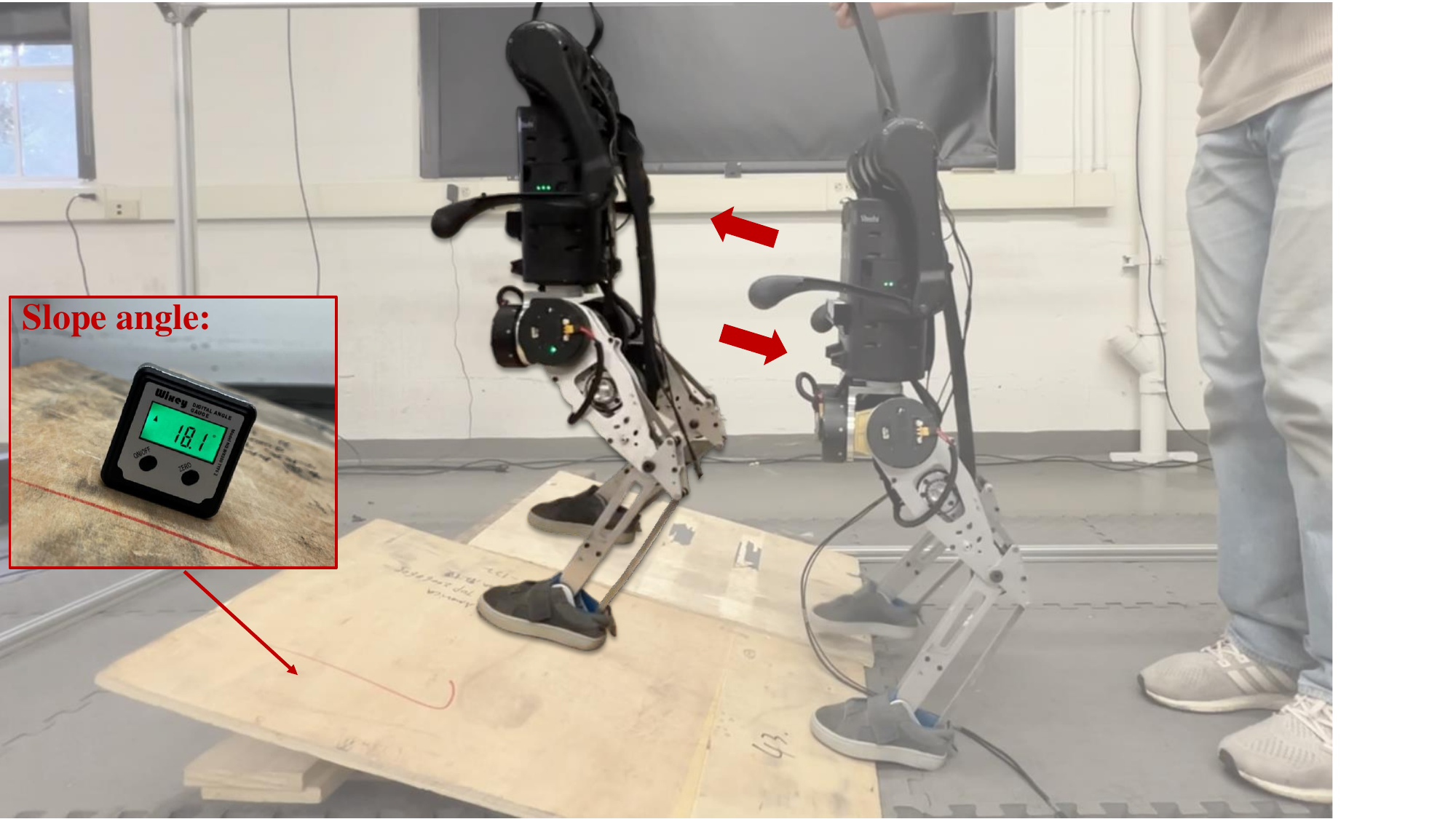}
         \caption{\centering}
         \label{fig:slope}
     \end{subfigure} 
     \begin{subfigure}[b]{0.40\textwidth}
         \centering
         \includegraphics[clip, trim=0cm 0cm 2.5cm 1cm, width=1\columnwidth]{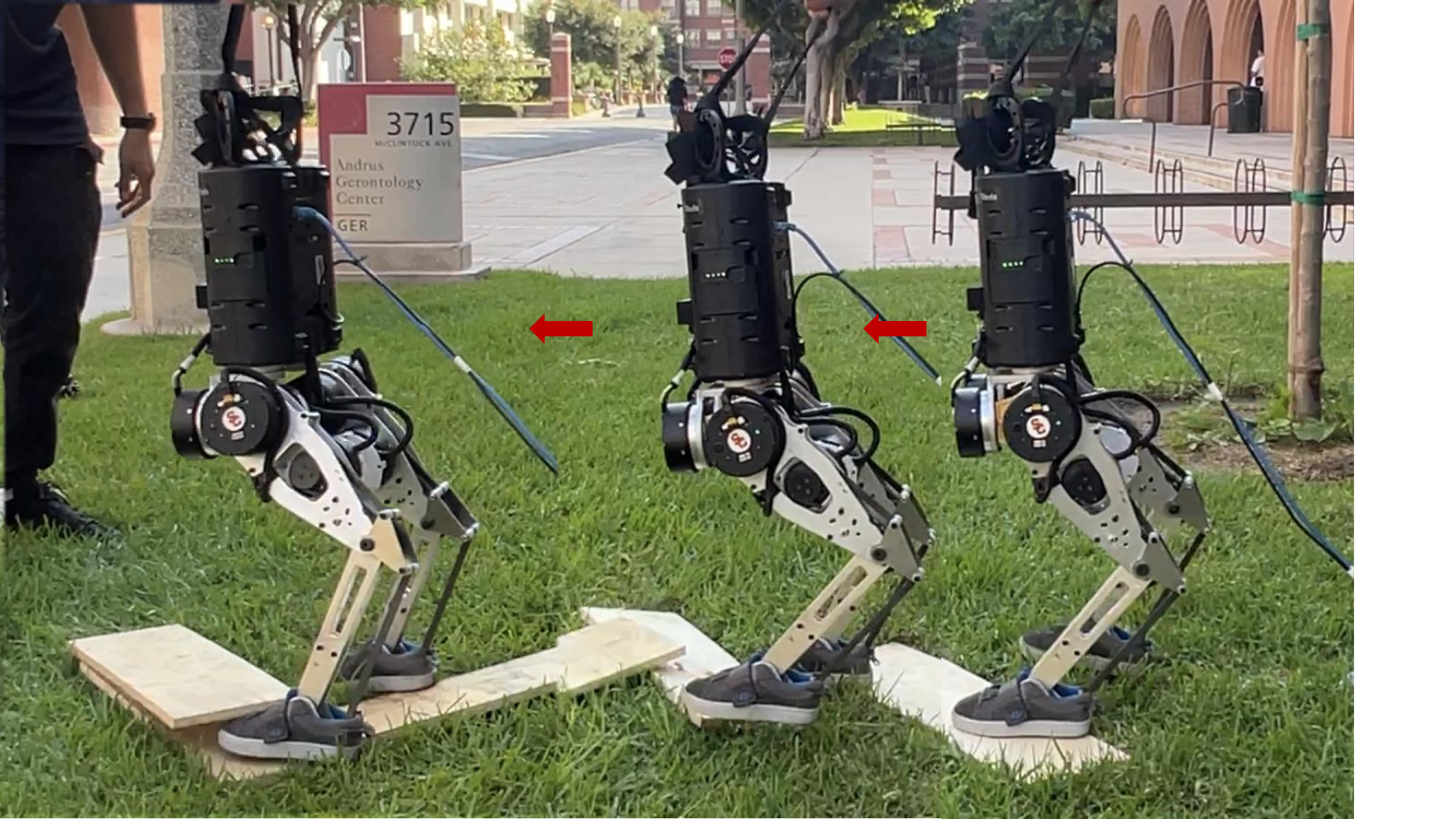}
         \caption{\centering}
         \label{fig:biped_grass}
     \end{subfigure} 
     \caption{ HECTOR experiment results: (a) Dynamically walking over random and unstable wood slats with 2.5 kg payload; (b) Balancing in-place while carrying 8 kg payload (50$\%$ robot mass); (c) Stepping in-place while carrying 4 kg payload; (d) Balancing when constantly moving the seesaw; (e) Dynamic turning with 2.5 kg payload (f) Walking up and down 18$^\circ$ slope; (g) Walking on grass with random wood slats. Full result video: \url{https://youtu.be/-r0QoxQgshk}}
     \label{fig:title}
     \vspace{-0.4cm}
\end{figure*}

Studying humanoid locomotion is an endeavor born out of a profound curiosity about understanding one of the most remarkable feats of nature: human movement. However, achieving dynamic locomotion and manipulation like humans in these robots remains a formidable challenge, necessitating revolutionary innovations in mechanical design, sensors, control, and artificial intelligence. 

A handful of exceptionally crafted humanoid hardware systems are currently available in the market, catering to a diverse range of business applications and research endeavors. These remarkable machines have showcased \jma{impressive} human-like locomotion and manipulation capabilities, exemplified by feats such as Boston Dynamics Atlas navigating a parkour course and Agility Robotics Digit transferring packages within warehouse settings. 
\rev{While often associated with high costs, these general-purpose robots possess immense future potential in aiding humans in exploration, disaster response missions, warehousing logistics, industrial inspections, and many more domains. These applications all require robust and dynamic control algorithms.}

\rev{In this study, we investigate the implementation of simplified-model-based control techniques on humanoid robots. Our goal is to validate the feasibility and efficiency of reduced-order models in dynamic humanoid loco-manipulation control. }


Moreover, we introduce a humanoid hardware and software solution, HECTOR project, aiming to advance and expand research in humanoid robotics by providing a robust testbed for both current and future control algorithms. This also includes the exploration of reinforcement learning (RL)-based control, a successful technique implemented in numerous quadrupedal robots for dynamic motions (e.g. \cite{hoeller2023anymal,cheng2023extreme, bellegarda2022robust}) but yet to be widely adopted on humanoid robots. 
HECTOR humanoid robot hardware will stand apart by being compact in scale, highly power-dense, cost-effective, and designed for easy operation and integration, addressing the accessibility and affordability barriers that currently hinder the advancement of humanoid robotics.


\subsection{Bipedal Locomotion Control}
Dynamic locomotion control has been a significant focal point in researchers' efforts to unlock the potential for human-like agility in bipedal locomotion. 
The primary advantage of \jma{bipedal} motion over other types of locomotion modes lies in its capability to navigate human-oriented terrains with uneven surfaces, various terrain types, and discontinuities.

Recent investigations on versatile bipedal locomotion control rely on simplified models such as the Zero Moment Point (ZMP), Spring-loaded Inverted Pendulum (SLIP), and Hybrid Zero Dynamics (HZD), all of which have been employed on locomotion controllers for bipedal robots (e.g. \cite{kajita2003biped, westervelt2003hybrid, hereid2014dynamic, sreenath2011compliant}). Notably, prior work in \cite{nguyen2018dynamic} has studied uneven discrete terrain bipedal locomotion through HZD-based control schemes and gait libraries. While this locomotion control framework excels at navigating discrete terrains, it requires precise terrain data for footstep planning.

Quadratic Programming (QP) based optimal control is another popular avenue in bipedal locomotion control. By linearizing the complex dynamics of the humanoid robots, an inverse dynamics problem can be efficiently solved by a QP solver (e.g. \cite{wensing2013generation,reher2020inverse,herzog2016momentum, xiong2018bipedal}). Authors in \cite{kim2020dynamic} realized stable and robust bipedal locomotion on humanoid robots with passive ankles through a time-to-velocity-reversal (TVR) planner and task-oriented Whole-body Locomotion Control (WBLC). The proposed framework is versatile and transferable across different robot platforms. \jma{However,} it requires the complex whole-body dynamics of the robot for obtaining task-space commands and solving inverse-dynamics QP.

With the aim of mitigating the requirement for precise terrain information and the complexity of whole-body dynamics derivation in the above bipedal locomotion control schemes, we have introduced a Force-and-moment-based MPC for dynamic and versatile bipedal locomotion behaviors in \cite{li2021force}, which is a middle ground that consists of reduced-order modeling, adaptivity to perturbations, and extendability to loco-manipulation.

Model Predictive Control (MPC) stands as a potent approach for achieving real-time planning and control, whether implemented with a reduced-order model or a nonlinear dynamics model. Recent implementations of MPC-based control in legged robots, such as the MIT Cheetah (\cite{kim2019highly,di2018dynamic}) and ANYmal quadruped (\cite{sleiman2021unified}), have demonstrated the viability of this scheme. However, in the realm of humanoid robots, the MPC-based approach has primarily functioned as a high-level planner for generating trajectories and planning footstep locations (e.g. \cite{scianca2020mpc,scianca2016intrinsically,brasseur2015robust}). A higher-frequency inverse-dynamics-based optimal low-level control has traditionally been employed together with MPC to manage locomotion and balance (e.g. \cite{chignoli2021humanoid, daneshmand2021variable}), utilizing the whole-body dynamics model (e.g., Whole-body Control in \cite{lee2016balancing}).

In our work, we rely on the proposed Force-and-Moment-Based MPC as the only primary optimal control layer, without the need for additional low-level WBC for enhancing balancing. To our knowledge, HECTOR represents the first instance of using only the simplified rigid body dynamics (SRBD)-based MPC to execute dynamic bipedal locomotion and loco-manipulation in hardware, thereby establishing a robust baseline for dynamic control and promising potential for further integration and development.

\subsection{Loco-manipulation on Legged Robots}
Control strategies for quadrupedal robots in loco-manipulation have achieved success in various domains, showcasing proficiency in robust object handling in \cite{rigo2023hierarchical}, non-prehensile manipulation in \cite{sleiman2021unified}, and dynamic load-bearing in \cite{sombolestan2023adaptive}.

Furthermore, there also has been a wide array of studies focused on multi-contact loco-manipulation on humanoid robots. Many of these studies have adopted a two-stage approach, which involves using high-level planner multi-contact preview and trajectory planning, and subsequently employing WBC to effectively track task-space commands such as in \cite{sentis2010compliant,audren2014model,murooka2021humanoid}. One notable challenge encountered when dealing with control in these scenarios is the accurate modeling of external objects' dynamics. As such, a simplified representation of load dynamics, incorporating the fundamental aspects of the object's dynamics, can be derived for model-based control. For example, object dynamics effects may be accounted for by external forces and wrenches applied to the robot, as seen in \cite{sleiman2021unified, agravante2019human}. 
In our approach, we simplified the object dynamics as the weight of the object applied to the robot SRBD model, as proposed in the author's prior work (\cite{li2023multi}). This approach incorporates the essential dynamic effects of the object, enabling the robot to compensate for the added mass and inertia within the SRBD model while maintaining linearity in the dynamics formulation.


\begin{figure}[!t]
    \center
    \includegraphics[clip, trim=-1cm 6cm 13cm 0cm, width=0.85\columnwidth]{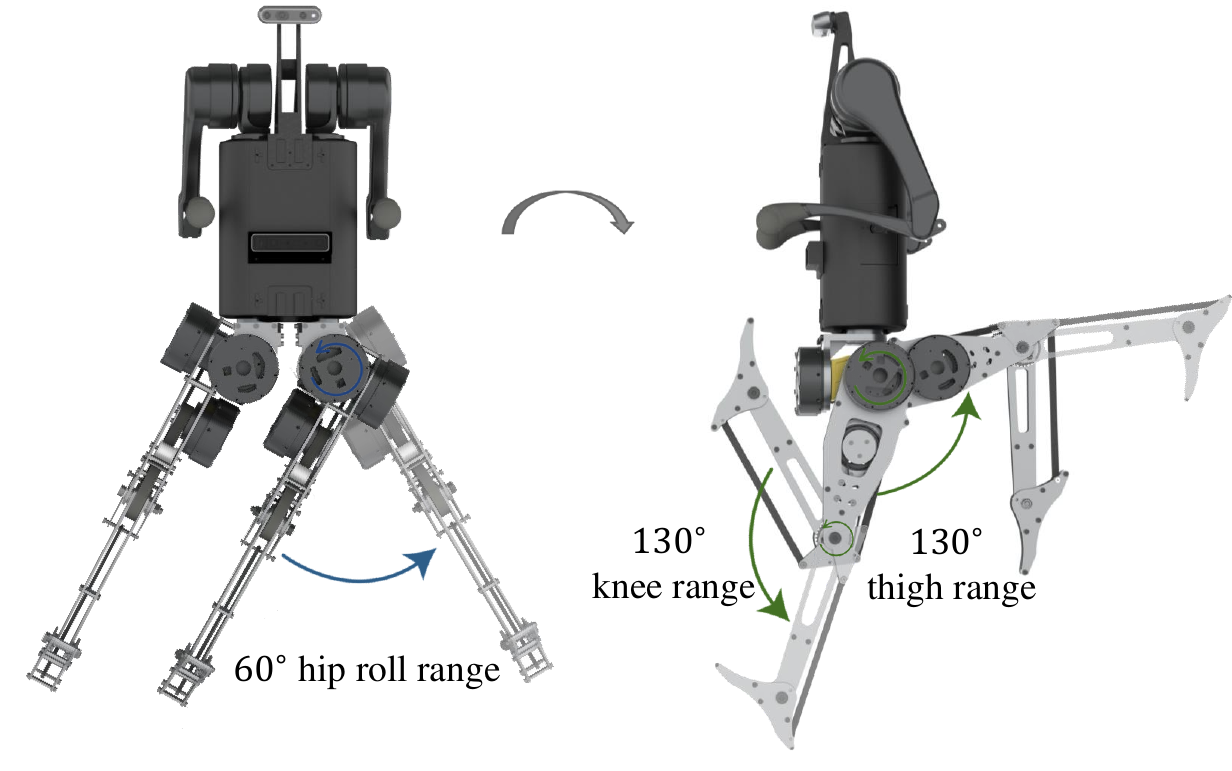}
    \caption{ HECTOR leg range of motion is illustrated by the range of motion of hip roll, thigh, and knee joints.  }
    \label{fig:range}
    \vspace{-0.3cm}
\end{figure}

\begin{figure*}[!t]
    \center
    \includegraphics[clip, trim=-1cm 8.5cm 0cm 0.6cm, width=2\columnwidth]{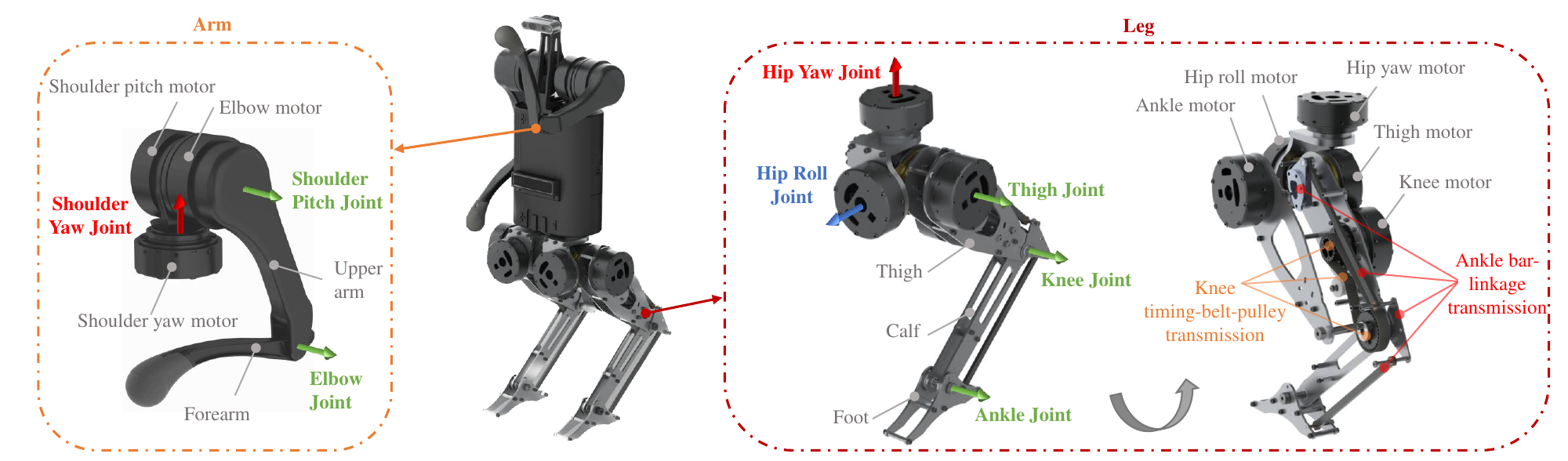}
    \caption{HECTOR's arm and leg kinematic definitions and exploded view of transmission systems.  }
    \label{fig:hardware}
    \vspace{-0.3cm}
\end{figure*}


\subsection{Humanoid Robot Hardware}
Humanoid control research frequently demands rigorous hardware validation, necessitating a robust and reliable humanoid testbed.
Traditional humanoid robots seen in 2000's and early 2010's such as ASIMO (\cite{asimo2002}), HUBO (\cite{park2005mechanial}), and TORO (\cite{englsberger2014overview}) rely on rigid actuators (i.e., high gear ratio harmonic drives) that excel in precise position control, yet this design limits their capacity for speed and impact mitigation - essential for dynamic motions. 

In contrast, WALK-MAN (\cite{tsagarakis2017walk}) and Agility Robotics' Digit utilize series-elastic actuators, offering improved impact absorption (\cite{abate2018mechanical}). 

On the other hand, Boston Dynamics's Atlas showcased unparalleled agility and complex terrain capabilities through hybrid electric and hydraulic actuators. However, this system's high cost and closed platform pose difficulties for mass adoption in scientific research. 

Lately, legged robot design methodology using back-drivable quasi-direct-drive (QDD) actuators has enabled affordable and high-performance dynamic quadruped robots that are capable of high-speed motion and high-impact mitigation (\cite{seok2014design}) such as MIT Cheetah 3 (\cite{bledt2018cheetah}) and mini Cheetah (\cite{katz2019mini}). However, this innovative approach, while prevalent in quadruped robots, has not yet seen widespread adoption in humanoid robots, presenting a unique opportunity.

Recent literature presented several small-scale torque-controlled bipedal and humanoid robots specifically designed to achieve dynamic and agile motions. Notably, Little HERMES (\cite{ramos2018facilitating}), Bolt (\cite{daneshmand2021variable}), MIT humanoid (\cite{chignoli2021humanoid}), , BRUCE (\cite{liu2022design}), and Tello leg (\cite{sim2022tello}) have showcased promising dynamic capabilities. 

In our work, we introduce HECTOR, a new ground in the field of small-scale humanoid robots with a unique approach to achieving dynamic motions. Leveraging QDD actuators allows HECTOR to excel at compliance and adaptivity in balancing and loco-manipulation and is best suited for force-based control frameworks.  This novel platform combines portability, modularity, high degrees of freedom, power density, and ease of operation, paving the way for a new generation of agile and versatile robots. This paper also provides an in-depth examination of HECTOR's dynamic locomotion and loco-manipulation abilities, offering concrete evidence of HECTOR's hardware capabilities.


\subsection{Main Contributions:}
This work builds upon the authors' prior conference papers on Force-and-moment MPC for humanoid locomotion and loco-manipulation (\cite{li2021force,li2023multi}). The additional contributions in this work, with respect to (w.r.t.) these papers, are as follows:
\begin{itemize}
    \item We present a generalized Force-and-moment-based MPC approach with simplified rigid body dynamics (SRBD) for generalizing dynamic locomotion and loco-manipulation control on humanoid robots. 
    
    \item To validate our proposed MPC framework, we have developed a portable, power-dense, cost-effective, and torque-controlled modular humanoid hardware, HECTOR, with a novel hybrid transmission system for achieving dynamic and agile motions. 
    
    \item W.r.t. prior papers, we have successfully demonstrated new and robust locomotion experiments on humanoid hardware. We have achieved stable, 3-D dynamic, and disturbance-resistant locomotion behavior with only an SRBD MPC as the primary control layer. 
    
    \item We have also successfully demonstrated dynamic loco-manipulation over uneven terrain on humanoid hardware, carrying a 2.5 kg load and walking over random and unstable wood piles.
    
    \item HECTOR simulation and controllers are made available as an open-sourced project for developing dynamic humanoid robots. 

\end{itemize}

The rest of the paper is organized as follows. Section \ref{sec:hardwareDesign} presents the design and hardware specifications of HECTOR. Section \ref{sec:sysArchi} is a brief overview of the system architecture of the robot and control system. Section \ref{sec:approach} starts by introducing the dynamics model of HECTOR and the process of simplification to SRBD and load dynamics. It follows by presenting the Force-and-moment-based MPC's formulation, simplification, low-level control, and swing leg policies for dynamic loco-manipulation. Lastly, we present the results of the proposed control schemes, validated both in numerical simulations and hardware experiments. 

\section{HECTOR Hardware Design}
\label{sec:hardwareDesign}

\begin{table}[t]
	\centering
	\caption{HECTOR Physical Properties}
	\begin{tabular}{ccc}
		\hline
		Property & Symbol & Value \\
		\hline
		Mass, $\unit{kg}$ & $m$  & 16.0  \\[.5ex]
		SRBD MoI, $\unit{kg}\cdot \unit{m}^2$  & $\prescript{}{\mathcal{B}}{I_{xx}}$  & 0.541 \\[.5ex]
		& $\prescript{}{\mathcal{B}}{I_{yy}}$ & 0.520  \\[.5ex]
		& $\prescript{}{\mathcal{B}}{I_{zz}}$ & 0.069  \\[.5ex]
		Upper/Lower Leg Length, $\unit{m}$ & $l_{leg} $ & 0.22  \\[.5ex]
		Foot Length, $\unit{m}$ & $l_{t}$ & 0.09 \\[.5ex]
		& $l_{h}$ & 0.05 \\[.5ex]
            Arm Length,  $\unit{m}$ & $l_{arm}$  & 0.20 \\ [.5ex]
		\hline 
		\label{tab:robotProperties}
	\end{tabular}
        \vspace{-0.3cm}
\end{table}

In our work, to validate our proposed control scheme, we present HECTOR, a torque-controlled humanoid testbed designed to perform highly dynamic behaviors in hardware experiments.  HECTOR is a versatile and robust small-scale humanoid platform, designed to offer accessible and cost-effective hardware solutions that are easy to operate, making it an ideal choice for research and development purposes.
Additionally, we offer open-source control and simulation software for HECTOR in both ROS and MATLAB. This initiative aims to expedite advancements in the field of humanoid robotics by providing a standard and available software platform.

\subsection{Mechanical Design}
HECTOR, as illustrated in Figure \ref{fig:hardware}, is primarily divided into three modular segments: the arms, trunk, and legs. This modular approach enables the robot to be easily reconfigured between bipedal and humanoid forms to suit various application needs. Standing at a height of 0.85 meters, HECTOR'S size resembles that of a child with physical properties summarized in the Table \ref{tab:robotProperties}. Each modular leg consists of 5 degrees of freedom including hip yaw, hip roll, thigh, knee, and ankle joints to achieve a large range of motion as seen in Figure \ref{fig:range}. In the design and development phase of HECTOR's leg structure, the primary focus was to strike a balance between maintainability, cost-effectiveness, and performance. The structural components of the legs are designed around the use of Aluminum 7075 T6 laser-cut plates and commercially available parts to offer rigidity for robust dynamic motions while maintaining a simplified design for cost-efficient prototyping as well as ease of serviceability.

Transitioning to the upper body, the arm module of HECTOR is designed to have a yaw joint at the shoulder and two pitch joints at the shoulder and elbow. This design, highlighted in Figure \ref{fig:hardware}, ensures that HECTOR remains lightweight yet capable of performing a diverse array of tasks.


\begin{figure*}[!t]
    \center
    \includegraphics[clip, trim=0cm 0.5cm 0cm 1cm, width=2\columnwidth]{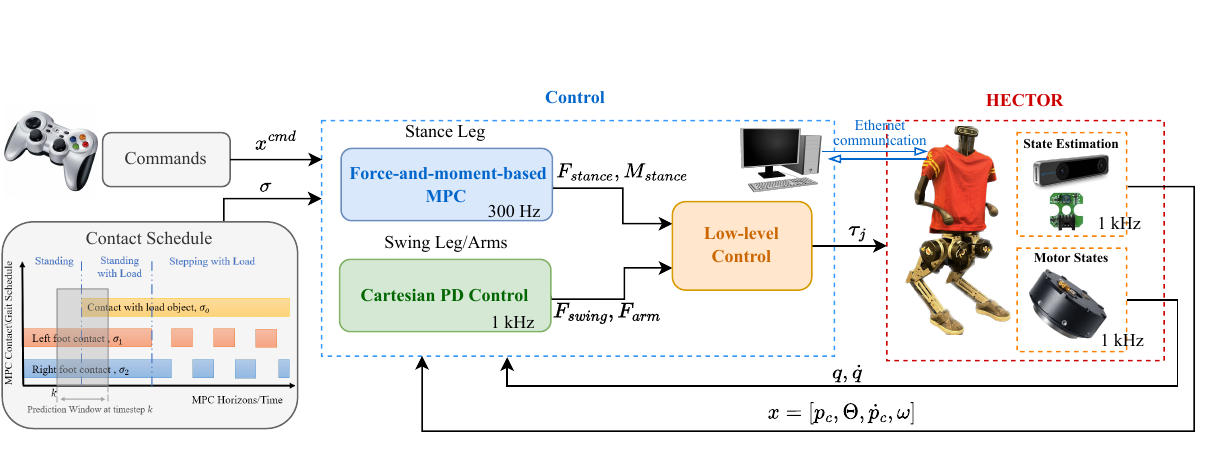}
    \caption{ System and Control Architecture of HECTOR.  }
    \label{fig:systemArchitecture}
    \vspace{-0.3cm}
\end{figure*}

\begin{table}[!t]
    \centering
    \caption{HECTOR Joint Configurations }
    \begin{tabular}{ccccc}
    \hline
	  Property & QDD Actuator & QDD + Timing Belt  \\
    \hline 
	Gear Ratio    & 9.1 &  14.1   \\ [.5ex]
        Peak Torque, $\unit{Nm}$     &  33.5 &51.9    \\ [.5ex]
        Peak Speed,  $\unit{rad/s}$  & 21.0   & 13.6     \\ [.5ex]

    \hline 
    \label{tab:Joint Actuator Specification}
    \end{tabular}
    \vspace{-0.5cm}
\end{table}

\subsection{Joint Actuation}

Building on HECTOR's compact and efficient design, the actuation system is engineered to fully exploit these attributes, enabling contact-rich, robust, and agile motions. 


To harness the advantages that quasi-direct-drive actuators provide, HECTOR joints are all driven by identical modular quasi-direct drive actuators equipped with a single-stage 9.1:1 planetary gearbox. This design ensures backdrivability and offers advantages in manufacturing simplicity, maintenance ease, and reduced costs. 

In an effort to minimize leg mass and concentrate body mass to the hip area, all motors have been positioned towards the upper thigh and hip regions. This increases the agility potential of the hardware platform while allowing for the possibility of approximating the robot dynamics as a rigid body model. Power is transmitted to the ankle joint via a 2-stage 4-bar linkage system, while the knee joint is powered through a timing-belt pulley system. The pulley system additionally serves to amplify the torque with a gear ratio of 1.55:1, thereby satisfying the torque requirements for various tasks without compromising agility. Therefore, two sets of torque and speed settings are implemented with HECTOR, both are summarized in Table \ref{tab:Joint Actuator Specification}.

\subsection{Hybrid Transsmision Systems}
While linkage systems are prevalent in robotics due to their rigidity and simplicity, they are unable to incorporate mechanical advantage without introducing nonlinearity that complicates control. On the other hand, timing belts are particularly well-suited for the increasing torque demands of bipedal and humanoid platforms, such as in \cite{ramos2019dynamic,chignoli2021humanoid}. However, they come with their own set of challenges, especially when it comes to maintaining compactness and robustness over multi-stage power transmissions across extended distances. 

To address this challenge, we've adopted a novel hybrid transmission mechanism that harnesses the strengths of both linkage systems and timing belts, achieving transmission efficiency, power density, and compactness. Specifically, our design integrates a two-stage linkage system forming two parallelograms, transmitting power from the ankle motor to the joint at a gear ratio of 1:1. Through the control-mechanical co-design process, where the increased knee torque requirement is observed to ensure dynamical capabilities according to simulation results in \cite{li2021force,li2023dynamic,li2023multi}, we have decided to use a timing belt-pulley system to both transmit and amplify power from the knee motor to the joint, achieving a gear ratio of 1.55:1. This design effectively amplifies the knee joint torque to 51.93 Nm, while retaining the necessary angular speed for dynamic motions. 

\section{System Architecture}
\label{sec:sysArchi}

In this section, the overall system and control architecture are presented. Figure \ref{fig:systemArchitecture} shows the block diagram of the system architecture. 
In a quick glance, we use a Logitech F710 wireless gamepad to send user commands to the controllers, which are executed in real time on an external PC. The computed torque commands generated by the controllers are subsequently sent to the robot's motor control boards. We retrieve information about the motor states and the robot states from motor feedback and state estimation. The communication between the external PC and the robot is facilitated through an Ethernet cable.

In detail, the robot states command $\bm x^{cmd}$ includes the desired velocities $\dot{p}_{c,x}^{cmd}, \dot{p}_{c,y}^{cmd}$ in the transverse plane and desired body yaw rate $\psi^{cmd}$, which are then mapped to a reference trajectory $\bm x^{ref} \in \mathbb{R}^{12\times h}$ with $h$ prediction horizons at each MPC time step for the controller to track. The reference trajectory at each horizon consists of robot CoM position $\bm p_c \in \mathbb{R}^3$, Euler angles $\bm \Theta  = [\phi, \theta, \psi]$, CoM velocity $\dot{\bm p}_c \in \mathbb{R}^3$, and angular velocity $\bm \omega \in \mathbb{R}^3$.

The contact schedule summarizes time-based periodic contact sequences based on the desired gait period length (typically in the range of 0.3 to 0.7 s for HECTOR) and loco-manipulation task. This gait sequence determines, at any time $t$, whether the leg is in stance phase (contact), or swing phase  (no contact). When leg $m$ is in stance phase, $m = 1, 2$, the ground reaction forces and moments for leg $m$ are then optimized by the MPC. When the leg is in swing phase, a Cartesian-space PD controller commands the foot to swing to a desired location for the next step. In the context of loco-manipulation tasks, HECTOR's arms are primarily employed for object manipulation and handling. Consequently, they are also controlled using a Cartesian PD controller. 

Stance foot control inputs $\bm u = [\bm F_{stance}; \bm M_{stance}]$, swing foot forces $\bm F_{swing}$, and arm forces $\bm F_{arm}$ are then mapped to corresponding joint torques $\bm \tau_j$ in a low-level controller by leg and arm Jacobian. Subsequently, the joint torque commands are sent to the motor control boards for executing the motion.
HECTOR uses a fused linear state estimator from the readings of onboard IMU and Intel T265 VIO, both of which are integrated onboard. The sensor fusion allows HECTOR to obtain accurate and high-frequency robot state feedback.   

\section{Proposed Approach}
\label{sec:approach}

This section will provide a comprehensive explanation of the proposed approach, including the dynamics modeling of the HECTOR humanoid robot, the SRBD assumptions, and the formulation of Force-and-moment-based MPC for dynamic locomotion and loco-manipulation.  

\subsection{Robot Dynamics Modelling}
\label{sec:dynamics}


\begin{figure*}[!t]
    \centering
    \includegraphics[clip, trim=0.5cm 9.5cm 0.5cm 0cm, width=2\columnwidth]{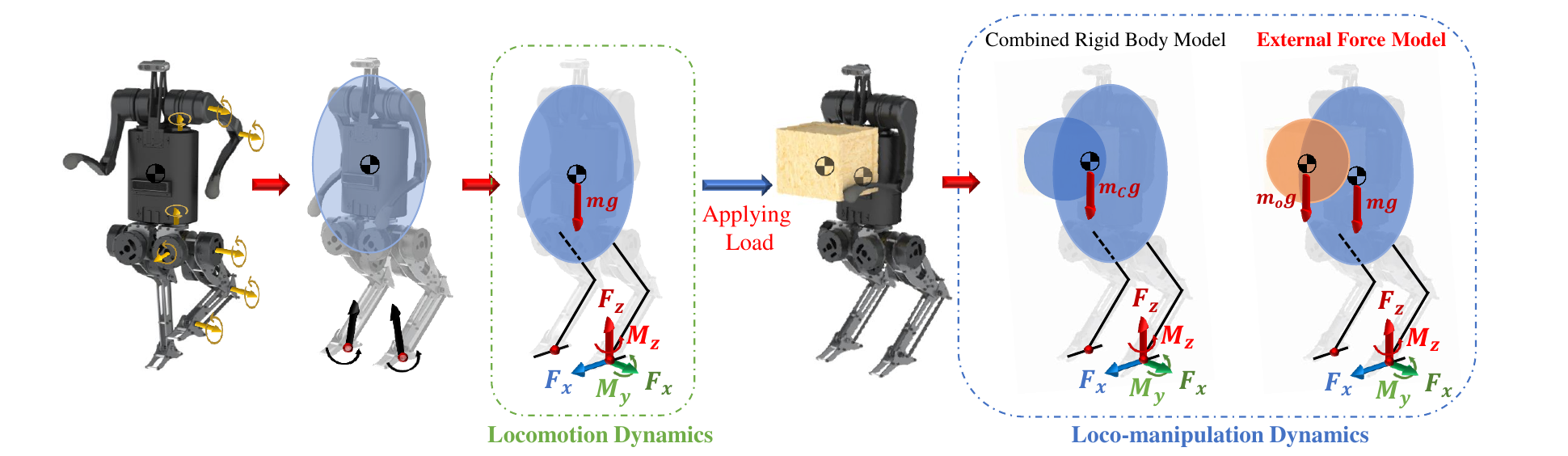}
    \caption{ Development of Humanoid SRBD in Locomotion and Loco-manipulation. }        
    \label{fig:dynamicsModel}
    \vspace{-0.3cm}
\end{figure*}

Firstly, we present the whole-body dynamics formulation of the HECTOR humanoid robot in the generalized coordinates. The joint-space generalized states ${\mathbf{q}} \in \mathbb{R}^{22}$ include $\bm p_c, \mathbf \Theta $, and $\bm q_j$.
\begin{align}
\label{eq:fullDynamics}
    \mathbf{H}(\mathbf{q})\ddot{\mathbf{q}} + \mathbf{C}(\mathbf{q}, \dot{\mathbf{q}}) = \bm \tau + \bm{J}(\mathbf{q})^\intercal \bm{\lambda} 
\end{align}

where $\mathbf{H} \in \mathbb{R}^{22 \times 22}$ is the mass-inertia matrix and $\mathbf{C} \in \mathbb{R}^{22}$ is the joint-space bias force. $\bm \tau$ represents the joint-space actuation. $\bm \lambda $ and $\bm{J}$ represent the external force applied to the system and corresponding Jacobian matrix.

\subsection{SRBD Model}
\subsubsection{Locomotion SRBD} \hfill

The motivation behind the utilization of reduced-order models for legged robots stems from the need for computational efficiency in online control and planning tasks, particularly in the context of predictive control (e.g., \cite{caron2016multi, bledt2017policy, xie2022glide}). In the prior work by the authors (\cite{li2021force}), three SRBD models for bipedal robots were introduced and analyzed. The aim was to identify an appropriate dynamics model for bipedal robots with 5 Degrees of Freedom (DoF) legs. It is concluded that 5-DoF leg configuration represented by HECTOR only requires 3-D ground reaction forces $[\prescript{}{\mathcal{F}}{F_x}; \prescript{}{\mathcal{F}}{F_y}; \prescript{}{\mathcal{F}}{F_z}]$ and 2-D ground reaction moments $[\prescript{}{\mathcal{F}}{M_y}; \prescript{}{\mathcal{F}}{M_z}]$, all in the foot frame. In MPC, to conveniently optimize for the control input in 3-D locomotion, we choose to represent the ground reaction forces and moments in the world frame for leg $m$ as $\bm F_m = [{F_x}; F_y; F_z]$ and $\bm M_m = [M_x; M_y; M_z]$.

Illustrated in Figure \ref{fig:dynamicsModel}, HECTOR's Locomotion SRBD encompasses the body, shoulders, hips, and upper thighs, all treated as a single combined rigid body. This combined rigid body captures the main dynamic effect of the system. The design of HECTOR is centered around the assumption of rigid body dynamics, whereby the mass of the remaining arm and leg components constitutes only 15$\%$ of the total robot mass. As a result, these components' inertia effects have been disregarded in the dynamics model. We assume the location of the ground reaction force and moment is a contact point on each foot, located at the projection of ankle joint location on the foot along the z-coordinate. The complete hardware actuation of the contact point, including 3-D linear movements, roll, and yaw, is fully achieved through the application of 5-D forces and moments.



\subsubsection{Loco-manipulation Dynamics} \hfill

Loco-manipulation and dynamic handling of heavy loads represent significant challenges in humanoid hardware and control. Typically, the dynamics of objects are not included in the humanoid dynamics model due to the variability in dynamics among different objects. Consequently, it becomes essential to capture the fundamental dynamics of the object to simplify and enable efficient and straightforward online utilization.

Our goal is to develop a generalized model for both locomotion and loco-manipulation. We expand the humanoid SRBD model to encompass dynamic object carrying and loco-manipulation by integrating the object dynamics into the robot's dynamics. In the author's previous work (\cite{li2023multi}), a comparison was made between two different formalisms of load dynamics within the SRBD framework:
\begin{itemize}
    \item \textit{Combined Rigid Body Model}: \rev{Treating the new SRBD as the combined rigid body of the robot and the object and updating the system dynamics w.r.t. the new combined CoM $m_c$;}
    \item \textit{External Force Model}: Treating the object dynamics as an external gravitational force/weight applied to the robot SRBD.
\end{itemize}

Upon examination, it was observed that the Combined Rigid Body Model exhibited significant state changes when transitioning from a no-load phase to a loaded phase. When an object is loaded on the robot, the dynamics model undergoes significant changes in parameters such as rigid body mass, inertia, CoM position, and Euler angles. This introduces instability that is particularly pronounced when dealing with heavy loads.

In this paper, we will delve into the utilization of the External Force Model, shown in Figure \ref{fig:dynamicsModel}, a formalism that relies only on the knowledge of the object's mass and the object's CoM relative to the robot's CoM. This approach aims to address the challenges associated with these state changes and enhance the robot's stability with heavy loads. 

Consequently, the simplified dynamics model for locomotion and loco-manipulation can be represented as
\begin{gather}
\label{eq:simpleDynamics1}
m (\ddot{\bm p}_c + \bm g) = m_o \bm g + \Sigma^2_{m = 1} \bm F_m  
\\
\label{eq:simpleDynamics2}
\frac{d}{dt}(\prescript{}{\mathcal{G}}{\bm I} \bm \omega) =
\bm r^{o} \times m_o \bm g + \Sigma^2_{m = 1} \{ \bm r_{m}^{f} \times \bm F_m + \bm M_m  \}
\end{gather}
where equation (\ref{eq:simpleDynamics1}) represents the force equilibrium in the world frame. The robot mass is denoted as $m$, the payload mass is denoted as $m_o$, and the gravity vector $\bm g$ is $[0; 0; g]$. Equation (\ref{eq:simpleDynamics2}) represents the moment equilibrium of the rigid body dynamics. $\prescript{}{\mathcal{G}}{\bm I}$ is the rigid body moment of inertia (MoI) in the world frame, which is obtained from the body frame MoI $\prescript{}{\mathcal{B}}{\bm I}$ and body rotation matrix $\mathbf R$, $\prescript{}{\mathcal{G}}{\bm I} = \mathbf R \prescript{}{\mathcal{B}}{\bm I} \mathbf R^T$. $\bm r^o$ represents the distance vector from robot CoM $\bm p_c$ to payload CoM $\bm p_{o}$ and $\bm r^f_m$ represents the distance vector from robot Co\jma{M} to $m$th foot contact point $\bm p_{f,m}$.

\subsection{Force-and-moment-based MPC}
\label{sec:MPC}

In the following sub-section, the proposed Force-and-moment-based MPC control framework is presented, including the selection of MPC states, assumptions to linearize the state-space dynamics, MPC formalism, and low-level control associated with MPC. 

SRBD-based Model Predictive Control (MPC) has been studied and demonstrated as a viable solution for governing dynamic locomotion in quadrupedal robots (e.g. \cite{kim2019highly, di2018dynamic}). In the author's prior work (\cite{li2021force}), a 2-D formalism of Force-and-Moment-Based MPC was introduced for bipedal locomotion in the sagittal and coronal planes. In this section, we present a developed Force-and-Moment-Based MPC framework for both 3-D locomotion and loco-manipulation.

MPC is a highly sought-after technique for high-level trajectory and footstep planning. In humanoid control, MPC is frequently combined with high-frequency WBC or TSC for instantaneous inverse-dynamics calculations. This often results in the control inputs from MPC having less influence on the system's stability.
However, in our work, we embrace a novel approach by allowing MPC to serve as our primary control scheme, fully harnessing the capabilities of the SRBD for dynamic locomotion and loco-manipulation.\hfill


\subsubsection{State-space Dynamics Linearization}\hfill

The robot state is defined as $\bm x = [\bm p_c; \bm \Theta; \dot{\bm p}_c; \bm \omega]$, and the derivative of the Euler angle and body angular velocity have the following relationship,
\begin{gather}
\nonumber
\omega_x = \dot{\phi}\sin{\theta} \sin{\psi} + \dot{\theta}\cos{\psi}, 
\\
\nonumber
\omega_y = \dot{\phi}\sin{\theta} \cos{\psi} - \dot{\theta}\sin{\psi}, 
\\
\omega_z = \dot{\phi}\cos{\theta} + \dot{\psi}.
\end{gather}
Equivalently, 
\begin{align}
\label{eq:omega2dthetadt}
\bm \omega = \underbrace{
\left[\begin{array}{ccc} 
\sin{\theta} \sin{\psi} & \cos{\psi} & 0 \\
\sin{\theta} \cos{\psi} & -\sin{\psi} & 0 \\
\cos{\theta} & 0 & 1  \end{array} \right] }_{\mathbf S_R}
\left[\begin{array}{c} \dot \phi \\ \dot \theta \\ \dot \psi \end{array} \right]. 
\end{align}

The left-hand side of equation (\ref{eq:simpleDynamics2}) may be approximated the following relation with the assumption of small angular velocity during locomotion,  
\begin{align}
\frac{d}{dt}(\prescript{}{\mathcal{G}}{\bm I} \bm \omega) = \prescript{}{\mathcal{G}}{\bm I} \dot {\bm \omega} + (\prescript{}{\mathcal{G}}{\bm I} \bm \omega) \approx \prescript{}{\mathcal{G}}{\bm I} \dot {\bm \omega},
\end{align}

With the above assumptions in linearization, the state-space equation of the rigid body dynamics model in Figure \ref{fig:dynamicsModel} can be written in the following form, 
\begin{gather}
\allowdisplaybreaks
\nonumber
\frac{d}{dt}
\left[\begin{array}{c} \bm p_c \\ \bm \Theta \\ \dot{\bm p}_c \\ \bm \omega  \end{array} \right] = \underbrace{
\left[\begin{array}{cccc} 
\mathbf{0}_3 & \mathbf{0}_3 & \mathbf{I}_3 & \mathbf{0}_3 \\
\mathbf{0}_3 & \mathbf{0}_3 & \mathbf{0}_3 & \mathbf {S}_R^{-1} \\
\mathbf{0}_3 & \mathbf{0}_3 & \mathbf{0}_3 & \mathbf{0}_3 \\
\mathbf{0}_3 & \mathbf{0}_3 & \mathbf{0}_3 & \mathbf{0}_3
\end{array} \right] }_{\mathbf A_c}
\left[\begin{array}{c} \bm p_c \\ \bm \Theta \\ \dot{\bm p}_c \\ \bm \omega  \end{array} \right] + \\
\setlength\arraycolsep{3pt}
\underbrace{
\left[\begin{array}{cccc} 
\mathbf{0}_3 & \mathbf{0}_3 & \mathbf{0}_3 & \mathbf{0}_3 \\
\mathbf{0}_3 & \mathbf{0}_3 & \mathbf{0}_3 & \mathbf{0}_3 \\ [3pt]
\mathbf{I}_3 \dfrac{\sigma_1}{m} & \mathbf{I}_3 \dfrac{\sigma_2}{m} & \mathbf{0}_3 & \mathbf{0}_3 \\ [6pt]
\dfrac{\sigma_1 [ \bm {r}^f_1 ]_\times}{\prescript{}{\mathcal{G}}{\bm I} }  & 
\dfrac{\sigma_2 [ \bm {r}^f_2 ]_\times}{\prescript{}{\mathcal{G}}{\bm I} } & \dfrac{\sigma_1}{\prescript{}{\mathcal{G}}{\bm I} } & \dfrac{\sigma_2}{\prescript{}{\mathcal{G}}{\bm I} }
\end{array} \right] }_{\mathbf B_c}
\underbrace{
\left[\begin{array}{c} \bm F_1 \\ \bm F_2 \\ \bm M_1 \\ \bm M_2  \end{array} \right] }_{\mathbf u} + 
\underbrace{
\left[\begin{array}{c} 
\mathbf{0}_{3\times1} \\
\mathbf{0}_{3\times1} \\[3pt]
\dfrac{m_o \bm g - m \bm g}{m}  \\[6pt]
\dfrac{\sigma_o \bm r^o \times m_o\bm g}{\prescript{}{\mathcal{G}}{\bm I}}
\end{array} \right] }_{\mathbf C_c}
\label{eq:stateSpace1}
\end{gather}
where $\mathbf I_3$ denotes 3-by-3 identity matrix. Note that the mapping between the rate of change of Euler angle and the body angular velocity $\mathbf S_R$ is not invertible at $\theta = \pm 90^\circ$. We do not intend to allow the robot to have such a configuration with this MPC. $\sigma_1$, $\sigma_2$, and $\sigma_o$ are Boolean variables that represent the time-based foot contact schedules and load contact schedules ($1$ for contact, $0$ for non-contact). 
The incorporation of contact schedules enables the MPC to possess information about changes in contact modes within the prediction horizons, illustrated by the contact schedule block in Figure \ref{fig:systemArchitecture}. This, in turn, facilitates improved optimization of the current control inputs, particularly with regard to potential under-actuation scenarios during contact mode changes.

With the nonlinearity $\mathbf {C}_c$ presented in equation (\ref{eq:stateSpace1}), we choose to modify robot state $\bm x$ and introduce the new MPC state variable selections $\mathbf x$ to include constant $1$, $\mathbf x = [\bm p_c; \bm \Theta; \dot{\bm p}_c; \bm \omega; 1]$. Consequently, the state-space dynamics equation is modified from equation (\ref{eq:stateSpace1}) with continuous matrices $\mathbf{A}_c$ and $\mathbf{B}_c$ rewritten as 
\begin{align}
    \frac{d\mathbf x}{dt} = \mathbf{A}_c(\mathbf{S}_R, \sigma_{1,2},\sigma_o,\bm r^o, m_o\bm g) \mathbf x + \mathbf{B}_c(\sigma_{1,2},\bm r^f_{1,2}) \mathbf u
\end{align}
and
\begin{align}
\mathbf {A}_c = 
\left[\begin{array}{cc} 
\mathbf{A}_c & \mathbf{C}_c \\
\mathbf{0}_{1\times12} & 0
\end{array} \right], \:
\mathbf {B}_c = \left[\begin{array}{c} \mathbf {B}_c \\ \mathbf{0}_{12\times1}
\end{array} \right] 
\end{align}

To utilize this linearized state-space dynamics equation within the MPC framework, we transform it into a discrete-time representation at the $k$th time step with MPC step $dt$: 
\begin{align}
\label{eq:discreteSS}
\mathbf {x}[k+1] = \mathbf {A}_k\mathbf x[k] + \mathbf {B}_k\mathbf u[k]
\\
\label{eq:discreteAB}
 \mathbf {A} = \mathbf I_{13} + \mathbf {A}_c dt ,\quad \mathbf {B} =  \mathbf {B}_c dt
\end{align}
\hfill

\subsubsection{MPC Formulation} \hfill

Now, we can establish the formal optimal control problem definition for the MPC that employs the linearized SRBD model. The convex MPC problem definition with a finite horizon $h$ is as follows
\begin{align}
\label{eq:MPCform}
\underset{\mathbf x,\mathbf u}{\operatorname{min}} \: \mathbf{\Sigma}_{k = 0}^{h-1} \: \: \underbrace{\vert \vert \mathbf x[k+1]-  \mathbf x^{ref}[k+1])\vert \vert^2_{\bm Q_k} }_{J_1} +  \underbrace{\vert \vert \mathbf{u} \vert \vert^2_ {\bm{R}_k}}_{J_2}
\end{align}

\begin{subequations}
\setlength\abovedisplayskip{-8pt}
\allowdisplaybreaks
\begin{gather}
\label{eq:dynamicCons}
\:\:\operatorname{s.t.} \quad \quad \mathbf {x}[k+1] = \mathbf {A}_k\mathbf x[k] + \mathbf {B}_k\mathbf u[k]  \\
\nonumber 
-\mu'  {F}_{m,z} \leq  F_{m,x} \leq \mu'  {F}_{m,z} \quad \quad\\
-\mu'  {F}_{m,z} \leq  F_{m,y} \leq \mu'  {F}_{m,z} \quad \quad
\label{eq:frictionCons}\\
\label{eq:forceCons}
0<  {F}_{min} \leq  F_{m,z} \leq  {F}_{max} \quad \quad\\
\label{eq:lfCons1}
[1, 0, 0]\cdot \mathbf R_{f,m}^T \bm M_m = \mathbf 0 \\
\nonumber
[0, 0, -l_t] \cdot \mathbf R_{f,m}^T \bm F_m \leq [0, 1, 0]\cdot \mathbf R_{f,m}^T \bm M_m \\
\quad \quad \leq [0, 0, l_h] \cdot \mathbf R_{f,m}^T \bm F_m 
\label{eq:lfCons2}
\\
\nonumber
[0, -l_t, -\mu' l_t]\cdot \mathbf R_{f,m}^T \bm F_m + [0, -\mu', \:\:\:1] \cdot \mathbf R_{f,m}^T \bm M_m \leq \mathbf{0} \\
\nonumber
[0, -l_t, \:\:\:\: \mu' l_t]\cdot \mathbf R_{f,m}^T \bm F_m + [0, \:\:\:\mu', \:\:\:1] \cdot \mathbf R_{f,m}^T \bm M_m \leq \mathbf{0} \\
\nonumber
[0, -l_h, -\mu' l_h]\cdot \mathbf R_{f,m}^T \bm F_m + [0, \:\:\:\mu', -1] \cdot \mathbf R_{f,m}^T \bm M_m \leq \mathbf{0} \\
\label{eq:lfCons3}
[0, -l_h, \:\:\:\mu' l_h]\cdot \mathbf R_{f,m}^T \bm F_m + [0, -\mu', -1] \cdot \mathbf R_{f,m}^T \bm M_m \leq \mathbf{0} 
\end{gather}
\end{subequations}

Equation (\ref{eq:MPCform}) describes the objectives of the convex MPC as a quadratic cost function. $J_1$ represents the cost of driving states to the desired trajectory based on the user commands. $J_2$ represents the cost of minimizing the control input to achieve better energy efficiency for locomotion. Both of the costs are weighted by the diagonal matrices $\bm Q_k$ and $\bm R_k$ at $k$th time step.

The optimal control problem is subjected to several constraints. Equation (\ref{eq:dynamicCons}) represents the discrete dynamics constraints derived from equation (\ref{eq:stateSpace1}-\ref{eq:discreteAB}). Equation (\ref{eq:frictionCons}) describes the contact point friction constraint, which follows an inscribed friction pyramid approximation, $\mu' = \sqrt{2}\mu/2$, for more conservative lateral foot forces, where $\mu$ is the contact friction coefficient of the ground. 

Equation (\ref{eq:lfCons1}-\ref{eq:lfCons3}) represents the line-foot dynamics constraints, where $\mathbf R_{f,m}^T$ denotes the foot rotation matrix w.r.t the world frame. Given the inherent challenges associated with modeling flat humanoid foot contact for humanoid robots (e.g., \cite{hauser2014fast,caron2015stability}), we have made the decision to simplify the representation of the foot contact on HECTOR to a line foot contact, which is also a common assumption on humanoid with 5-DoF legs (e.g., \cite{ahn2021versatile, ding2022orientation}). We modified the Contact Wrench Cone (CWC) introduced in \cite{caron2015stability} to enforce:
\begin{itemize}
    \item The x-direction ground reaction moment is zero, shown in equation (\ref{eq:lfCons1}),
    \item Prevention of toe/heel lift due to the y-direction ground reaction moment, shown in equation (\ref{eq:lfCons2}), 
    \item Friction constraints for y-direction reaction forces at line-foot heel and toe locations, shown in equation (\ref{eq:lfCons3}).
\end{itemize}
\hfill

\subsubsection{Condensing MPC-QP Problem for Real-time Computation}\hfill

Despite the extended prediction horizon embedded in the proposed MPC, resulting in large matrix sizes that demand significant computational resources for online optimal control problems, we can still optimize the convex MPC formulation to ensure efficient solutions as a Quadratic Programming (QP) optimization (e.g. \cite{wang2009fast, jerez2011condensed}).

In a standard non-condense MPC-QP problem definition, the augmented state $\mathbf X = \bigr[ \mathbf{x}[1] \dots \mathbf{x}[h];  \mathbf{u} [0] \dots \mathbf{u} [h-1] \bigr] $ consists of both states $\mathbf{x}$ and decision variables $\mathbf{u}$. A QP problem definition is,
\begin{gather}
    \underset{\mathbf X}{\operatorname{min}} \: \frac{1}{2}\mathbf X^T \mathbf H \mathbf X + \mathbf f^T \mathbf X
\\
\allowdisplaybreaks
\label{eq:qp2}
\:\:\operatorname{s.t.} \quad \mathbf A_{eq} \mathbf X = \mathbf b_{eq} \\
\label{eq:qp3}
\quad \quad \quad \: \mathbf C \mathbf X < \mathbf d
\end{gather}

We can express the problem (eqn. (\ref{eq:MPCform})) with dynamics constraints (eqn. (\ref{eq:dynamicCons})) in terms of the above standard non-condensed QP form according to \cite{wright1996applying}, where
\begin{gather}
\allowdisplaybreaks
\label{eq:h1}
\mathbf {H} = 
\text{blkdiag}(\{\bm Q_i \}^{h-1}_{i = 0},\{\bm R_i\}^{h-1}_{i = 0}), \\
\label{eq:f1}
\mathbf {f} = -\bigr[\{{\bm Q_i\mathbf x^{ref}_i}^T \}^{h-1}_{i = 0}, \{ {\mathbf 0_{12\times1}}\}^{h-1}_{j = 0}\bigr]^T;
\end{gather}
And
\begin{gather}
\nonumber
\setlength\arraycolsep{3pt}
\mathbf A_{eq} = 
\left[\begin{array}{cccccccc} 
\mathbf{I} &  &  &  & -\mathbf{B}_0 &  &  &\\
-\mathbf{A}_1 & \ddots &  &  &  & \ddots &  & \\
 & \ddots & \ddots &  &  &  & \ddots &  \\
 & & -\mathbf{A}_{h-1} & \mathbf{I} &  &  &  & -\mathbf{B}_{h-1}
\end{array} \right],
\\
\mathbf b_{eq} = \bigr[\mathbf A_0 \mathbf{X}[0], \mathbf{0}, \dots, \mathbf{0}\bigr]^T.
\end{gather}

\begin{table}[!t]
\vspace{0.2cm}
    \centering
    \caption{Average MPC Total Solve Time$^*$ Comparisons}
    \begin{tabular}{ccccc}
    \hline
	 Solver$\backslash$Approach  & Non-condensed & \textbf{Condensed}  \\
    \hline 
    \textbf{qpOASES (C++)} & & \\[.5ex]
	Standing, $\unit{ms}$    & 19.9 &  \textbf{3.2}   \\ [.5ex]
        Walking, $\unit{ms}$    &  12.9 &  \textbf{1.3}    \\ [.5ex]
    \hline 
    \textbf{quadprog (MATLAB)} & & \\[.5ex]
	Standing, $\unit{ms}$    & 432.5 &  \textbf{99.3}   \\ [.5ex]
        Walking, $\unit{ms}$    &  367.2 &  \textbf{24.9}    \\ [.5ex]
    \hline 
    \label{tab:solveTime}
    \end{tabular}
    \text{$^*$PC with AMD Ryzen 5600X @ 4.2 GHz on Ubuntu 20.04}
    \vspace{-0.4cm}
\end{table}

However, due to the considerably large size of the equality constraint matrix $\mathbf{A}_{eq} \in \mathbb{R}^{13h\times25h}, \mathbf{b}_{eq} \in \mathbb{R}^{13h}$, it is computationally expensive to solve for the solutions online. In addition, the decision variables we need for actuating the humanoid robot are only the current step control inputs $\mathbf{u}[0]$. Therefore, the resulting full-state calculation from the non-condensed approach can be unnecessary. Given that the state-space matrices $\mathbf{A}_d$ and $\mathbf{B}_d$ are time-varying and periodic, it is reasonable to apply a condensed and sparse QP formulation to solve the MPC problem according to \cite{jerez2011condensed}. A comparison table of the MPC solve time between non-condensed and condensed formulations on HECTOR hardware is provided in Table \ref{tab:solveTime}. The condensed approach for solving the MPC-QP problem allows us to run the hardware MPC very efficiently online with up to 1.5$\unit{s}$ prediction horizon and hyper-sample it up to 300$\unit{Hz}$.

\begin{figure*}[!t]
\vspace{0.2cm}
    \center
    \begin{subfigure}[b]{0.60\textwidth}
    \centering
    \includegraphics[width=1\textwidth]{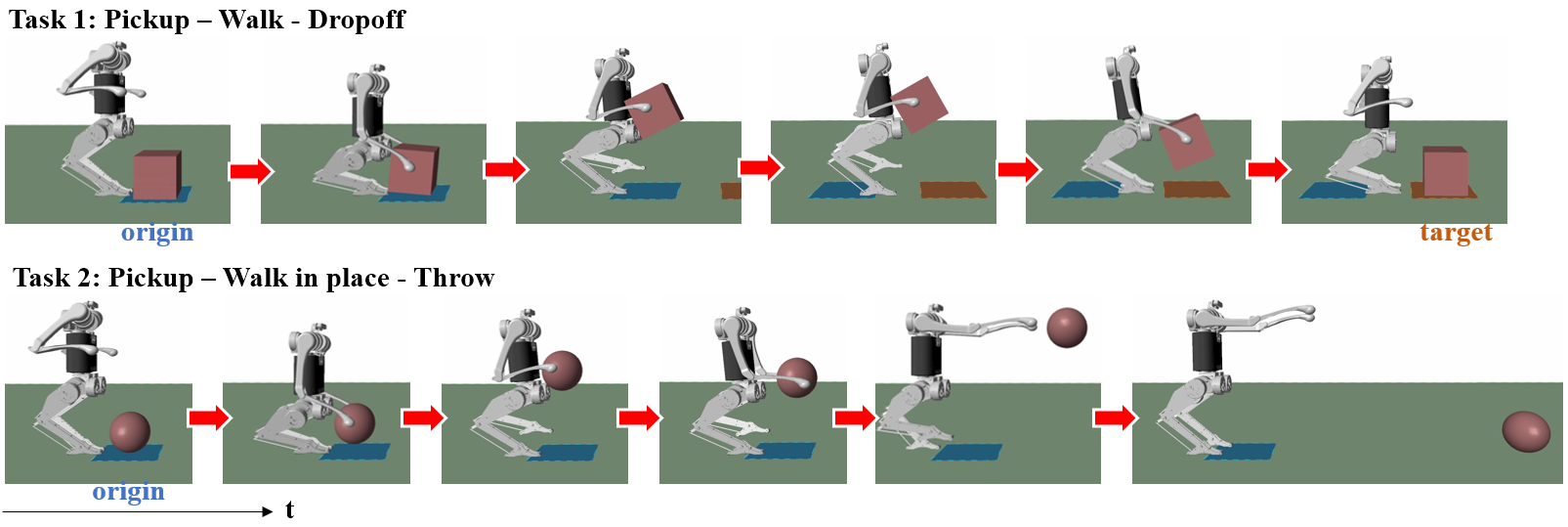}
    \caption{\centering}
    \label{fig:LocoSimSnaps}
    \end{subfigure}
    \begin{subfigure}[b]{0.39\textwidth}
    \centering
    \includegraphics[clip, trim=0cm 8cm 6.2cm 0cm, width=1\columnwidth]{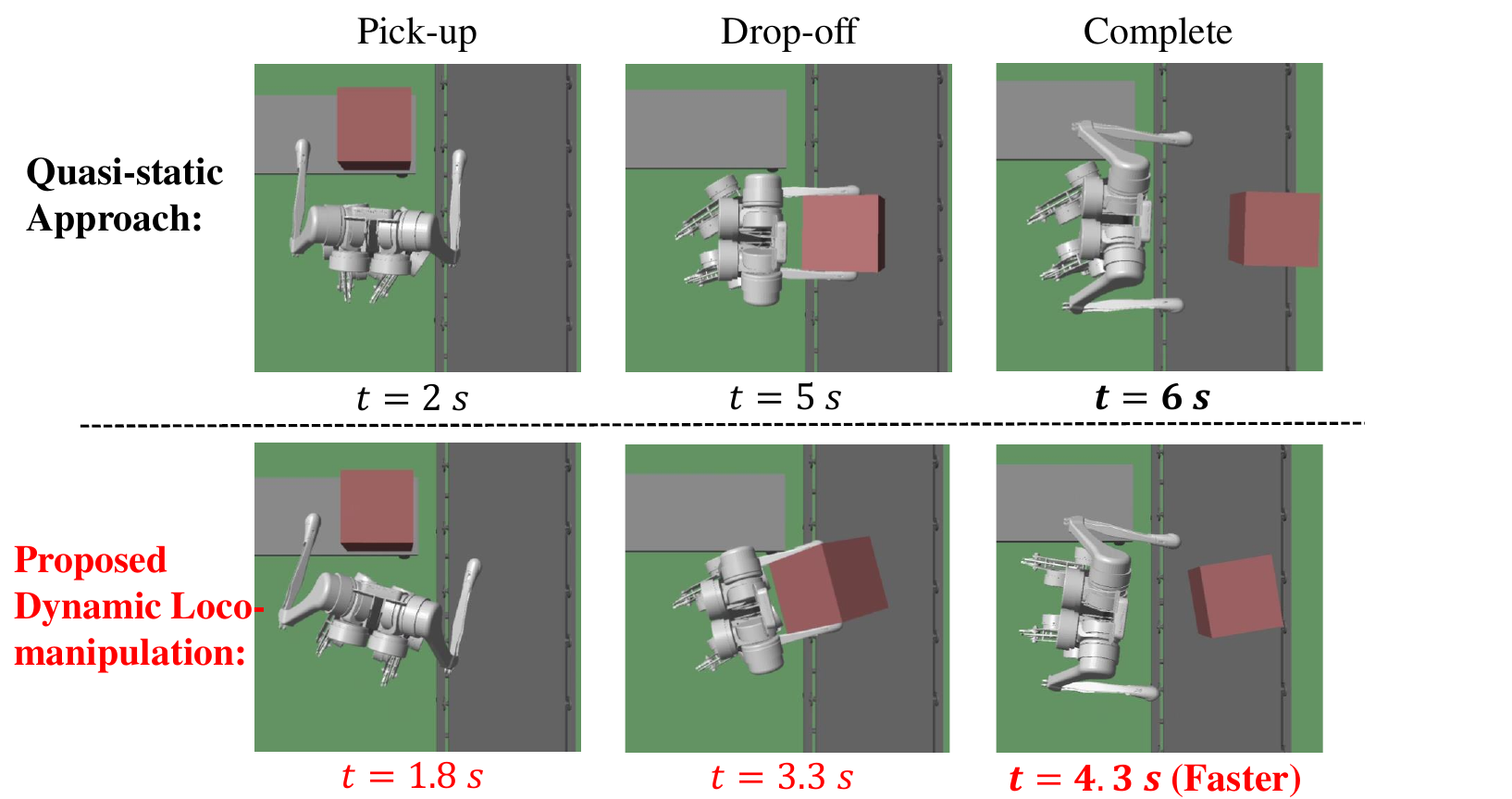}
    \caption{\centering}
    \label{fig:90turnComparison}
    \end{subfigure}
    \vspace{-0.1cm}
    \caption{ HECTOR loco-manipulation simulation snapshots: (a) Multi-contact loco-manipulation control including 1) Transferring a 4kg package and 2) dynamically throwing a 2kg ball while staying balanced; (b) Comparison of approaches in 90-degree turn package transferring. }
    \label{fig:snapshots}
    \vspace{-0.3cm}
\end{figure*}

   \begin{figure}[!t]
\vspace{0.2cm}
     \centering
     \begin{subfigure}[b]{0.5\textwidth}
         \centering
         \includegraphics[clip, trim=0cm 8.5cm 9cm 0cm, width=1\columnwidth]{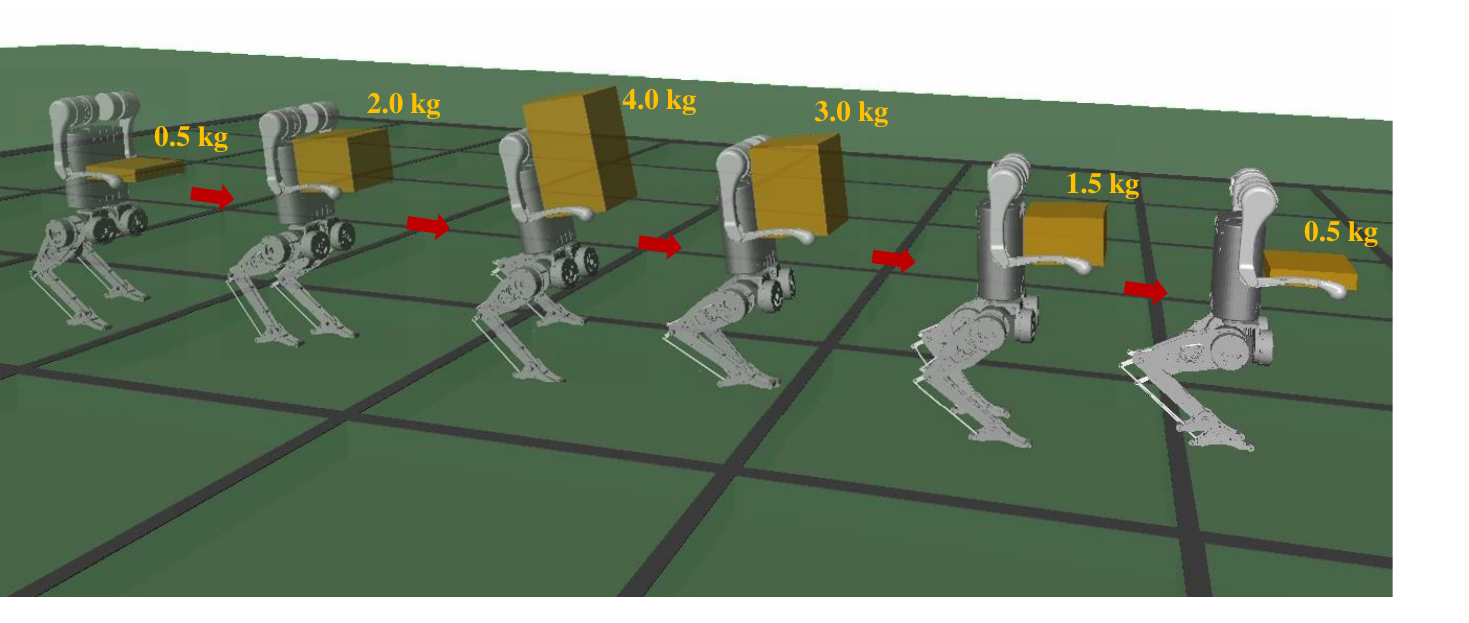}
         \caption{\centering}
         \label{fig:loco_snap}
     \end{subfigure} 
     \vspace{-0.2cm}
     \\
     \begin{subfigure}[b]{0.48\textwidth}
         \centering
         \includegraphics[clip, trim=3cm 9cm 3cm 9cm, width=1\columnwidth]{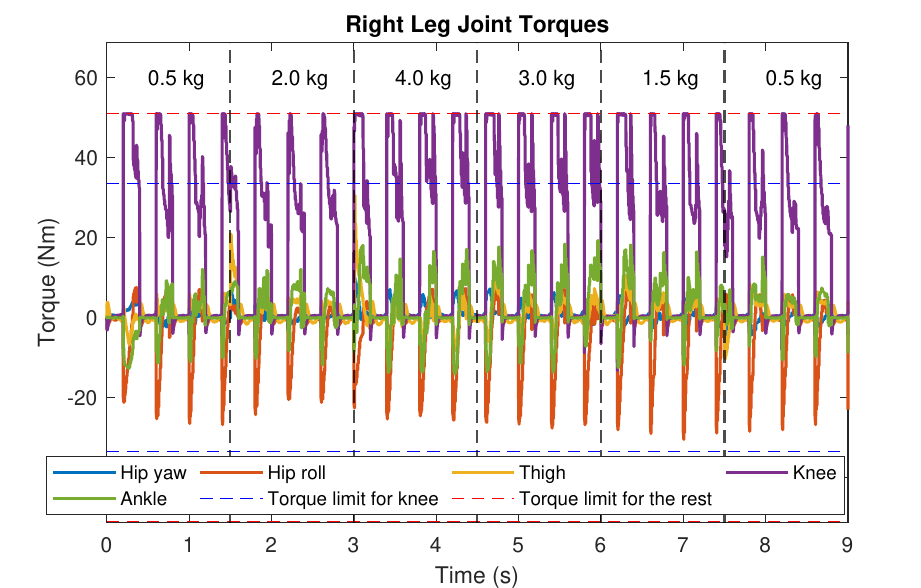}
         \caption{\centering}
         \label{fig:loco_torque}
     \end{subfigure}
     \vspace{-0.1cm}
        \caption{ HECTOR loco-manipulation with time-varying load: (a) Simulation snapshots; (b) Joint torque plots.} 
        \label{fig:loco}
        \vspace{-0.2cm}
\end{figure}

We apply the following formula to translate the $\mathbf{H}$ and $\mathbf{f}$ matrices from equation (\ref{eq:h1}-\ref{eq:f1}) to a condense and sparse QP form to solve efficiently,
\begin{gather}
\label{eq:h2}
\mathbf{H}^{cs} = 2( {\mathbf B_{qp}}^T\mathbf Q {\mathbf B_{qp}}+\mathbf R), \\
\label{eq:2}
\mathbf{f}^{cs} = 2 {\mathbf B_{qp}}^T\mathbf Q ({\mathbf A_{qp}}\mathbf x[0]-\mathbf x_{qp}).
\end{gather}
where $\mathbf x_{qp},\: \mathbf Q$, and $\mathbf{R}$ are all vertical concatenations of corresponding matrices from MPC time step 0 (current instance) to $h-1$. And,
\begin{gather}
\allowdisplaybreaks
\mathbf A_{qp} = [\mathbf{A}_0,\: \mathbf{A}_0\mathbf{A}_1, \: \dots \:,  \mathbf{A}_0\mathbf{A}_1\dots \mathbf{A}_{h-1}]^T, \\
\setlength\arraycolsep{2pt}
\mathbf B_{qp} = 
\left[\begin{array}{cccccc} 
\mathbf I & & & & &\\ [2pt]
\mathbf{A}_1 & \mathbf{I} & & & & \\[2pt]
\vdots & \mathbf{A}_{2} & \ddots&  & &\\[2pt]
\mathbf{A}_{h-3}^{h-3} & \vdots & \ddots & \ddots& & \\[4pt]
\mathbf{A}_{h-2}^{h-2} & \mathbf{A}_{h-1}^{h-3} & \dots & \mathbf{A}_{h-2} & \mathbf I & \\[4pt]
\mathbf{A}_{h-1}^{h-1} & \mathbf{A}_{h-1}^{h-2} & \mathbf{A}_{h-1}^{h-3} & \dots &\mathbf{A}_{h-1} & \mathbf{I}
\end{array} \right] \cdot
\mathbf{B}^{blkdiag}_{0\rightarrow h-1}
\end{gather}

\subsection{Low-level Control}

In contrast to many other strategies for bipedal locomotion control, which employ a low-level WBC with task-space commands as a high-frequency balancing controller to manage whole-body dynamics, we thoroughly investigate the effectiveness of Force-and-moment-based MPC as the primary locomotion controller. In our approach, we directly utilize the control input from MPC and map it to joint torques using contact Jacobian, with the assumption of the SRBD model and negligible leg dynamics. Hence, the $m$th leg joint torque $\bm \tau_m$ commands are
\begin{align}
\bm \tau_m = \left[\begin{array}{c} 
\bm J_v\\
\bm J_\omega
\end{array} \right]^T
\left[\begin{array}{c} 
\mathbf R_{f,m}^T (\bm F_m + \bm F_{swing,m}) \\
\mathbf R_{f,m}^T \bm M_m
\end{array} \right],
\end{align}
where $\bm J_v \in \mathbb{R}^{3\times 5}$ and $\bm J_\omega \in \mathbb{R}^{3\times 5}$ are linear and angular velocity parts of foot contact point Jacobian. Note that swing foot force $\bm F_{swing,m}$ and stance foot force cannot be both nonzero at any instance, and the foot is either under swing or stance phase according to the gait scheduler. The swing force is acquired by Cartesian-space PD control law driving the swing foot to the desired heuristic foot position $\bm p_m^{des}$ in \cite{raibert1986legged} and \cite{kim2019highly} and velocity $\dot{\bm p}_m^{des}$, 
\begin{gather}
    \bm p_m^{des} = \bm p_c + \frac{\dot{\bm p}_c\Delta t}{2} + k_c(\dot{\bm p}_c - \dot{\bm p}_c^{cmd}), \\
    \dot{\bm p}_m^{des} = 2\dot{\bm p}_c
\end{gather}
$\Delta t$ denotes the time duration of each foot swing and scalar $k_c$ represents the velocity feedback gain to achieve the desired walking speed. 

While holding objects with arms, the upper body joint commands are also calculated by the Cartesian space control and contact (hand) Jacobian with stiff PD gains. 



\section{Results}
\label{sec:Results}


 In this section, we will present the results of both numerical and hardware experiments to validate our proposed Force-and-moment-based MPC approach for dynamic humanoid control. We strongly encourage readers to view the full result videos\footnote{\url{https://youtu.be/-r0QoxQgshk}} that accompany this paper for visual aid and a comprehensive understanding of our work.
 
\begin{figure*}[!t]
\vspace{0.2cm}
     \centering
     \begin{subfigure}[b]{0.31\textwidth}
         \centering
         \includegraphics[clip, trim=1cm 2cm 2cm 0cm, width=1\columnwidth]{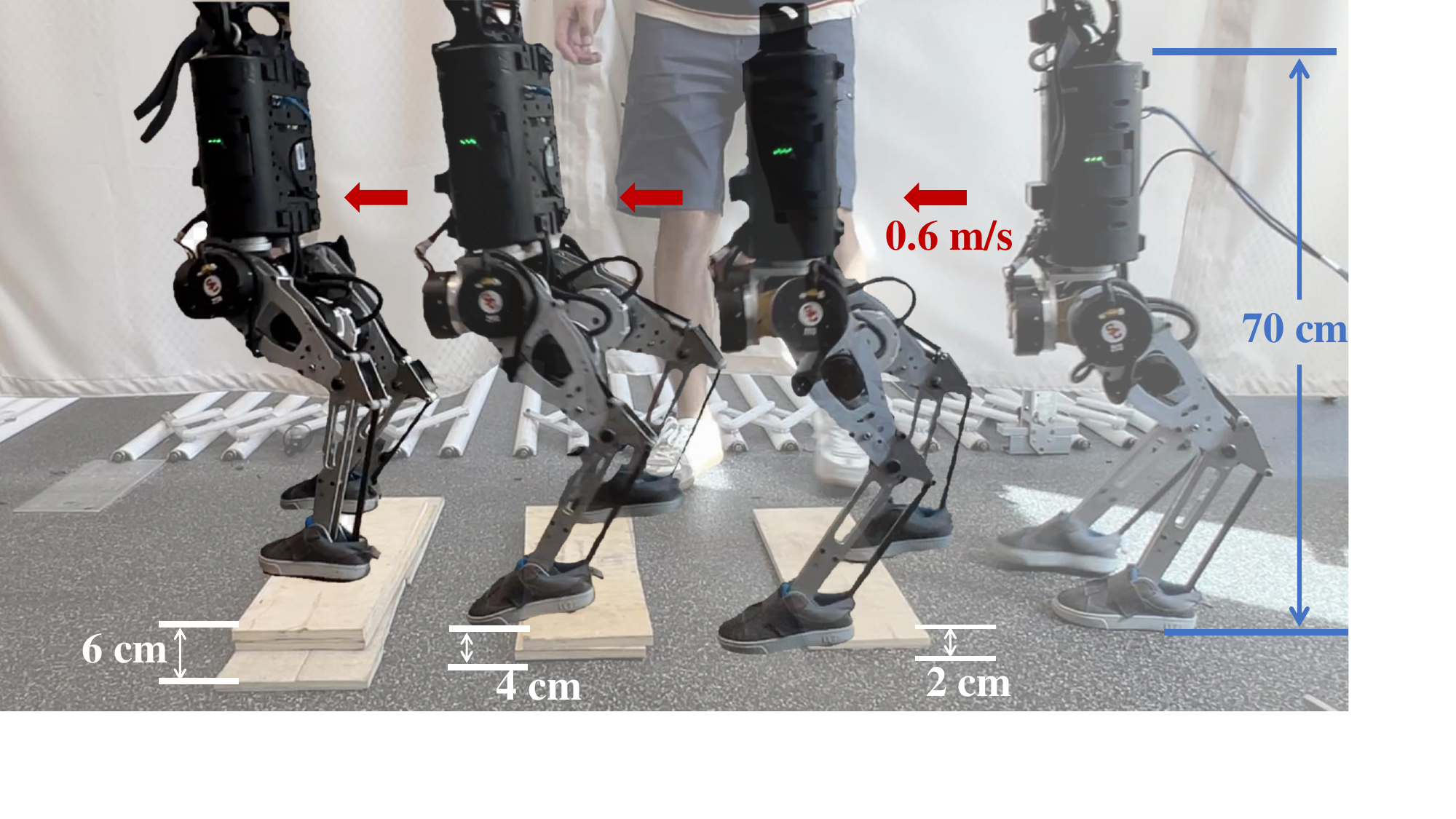}
         \caption{\centering}
         \label{fig:biped_terrain1}
     \end{subfigure} 
     \:\:
     \begin{subfigure}[b]{0.33\textwidth}
         \centering
         \includegraphics[clip, trim=1cm 1cm 0cm 1cm, width=1\columnwidth]{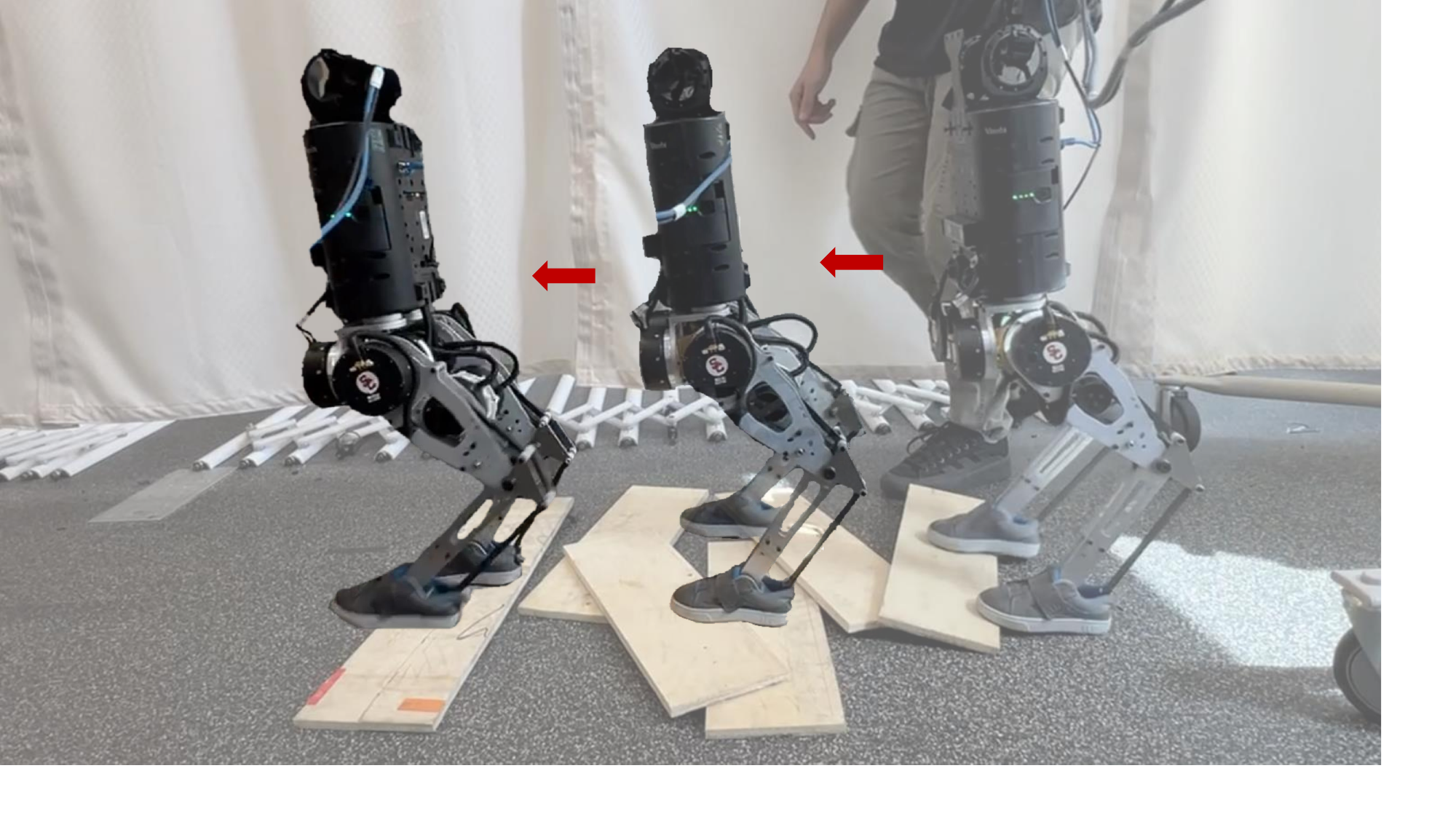}
         \caption{\centering}
         \label{fig:biped_terrain2}
     \end{subfigure} 
     \begin{subfigure}[b]{0.31\textwidth}
         \centering
         \includegraphics[clip, trim=1cm 2cm 2cm 0cm, width=1\columnwidth]{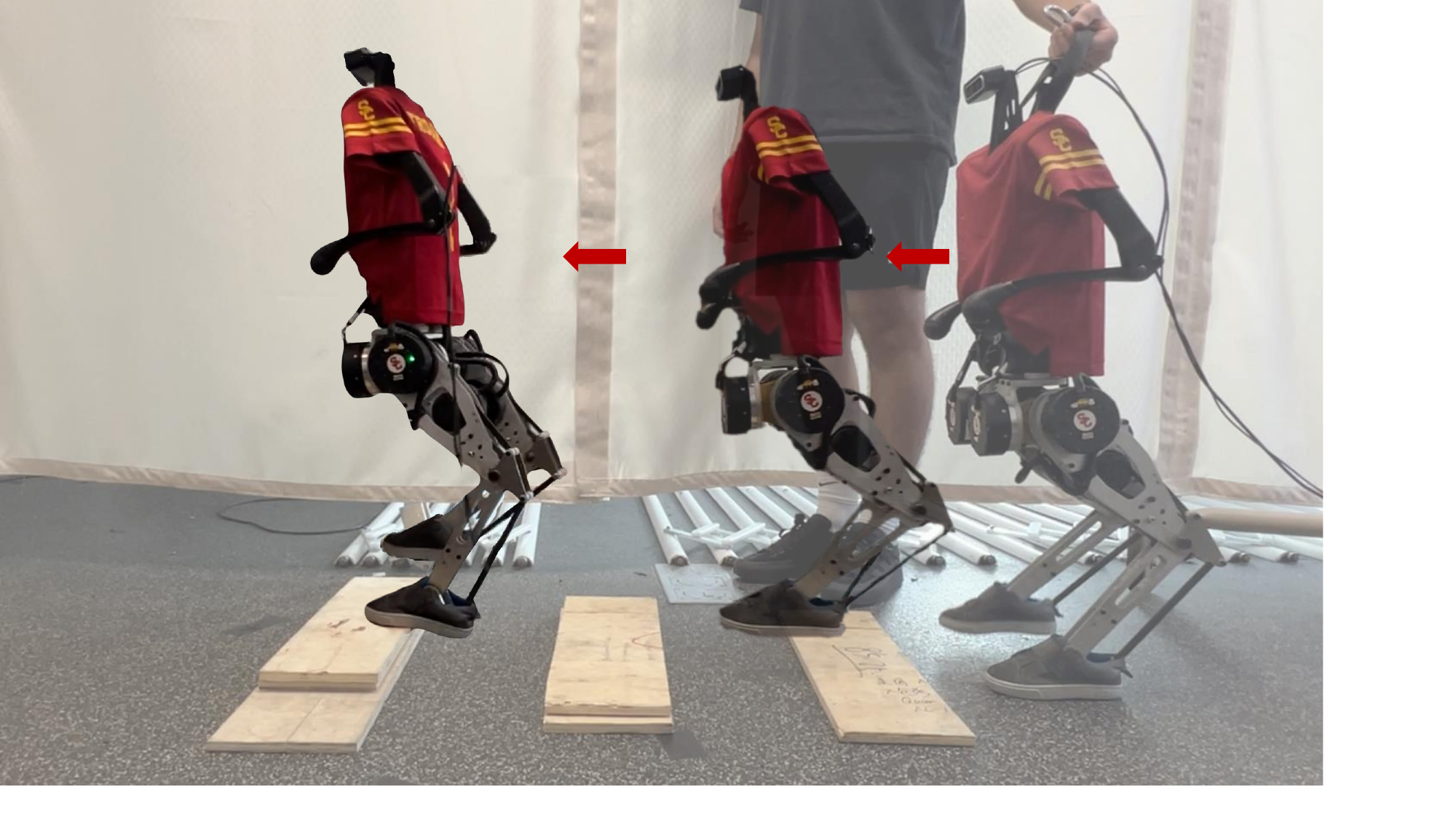}
         \caption{\centering}
         \label{fig:humanoid_terrain1}
     \end{subfigure} 
     \quad
     \vspace{-0.2cm}
        \caption{ HECTOR uneven terrain locomotion experiment snapshots: (a)  Walking over stacked wood slats with 0.6 m/s forward speed; (b) Walking over random wood slats; (c) Walking over stacked wood slats in humanoid form. Note all wood slat terrains are unstable. }
        \label{fig:expSnaps}
        \vspace{-0.2cm}
\end{figure*}

\begin{figure}[!t]
    \centering
    \begin{subfigure}[b]{0.46\textwidth}
    \includegraphics[clip, trim=0cm 3cm 3cm 1cm, width=1\columnwidth]{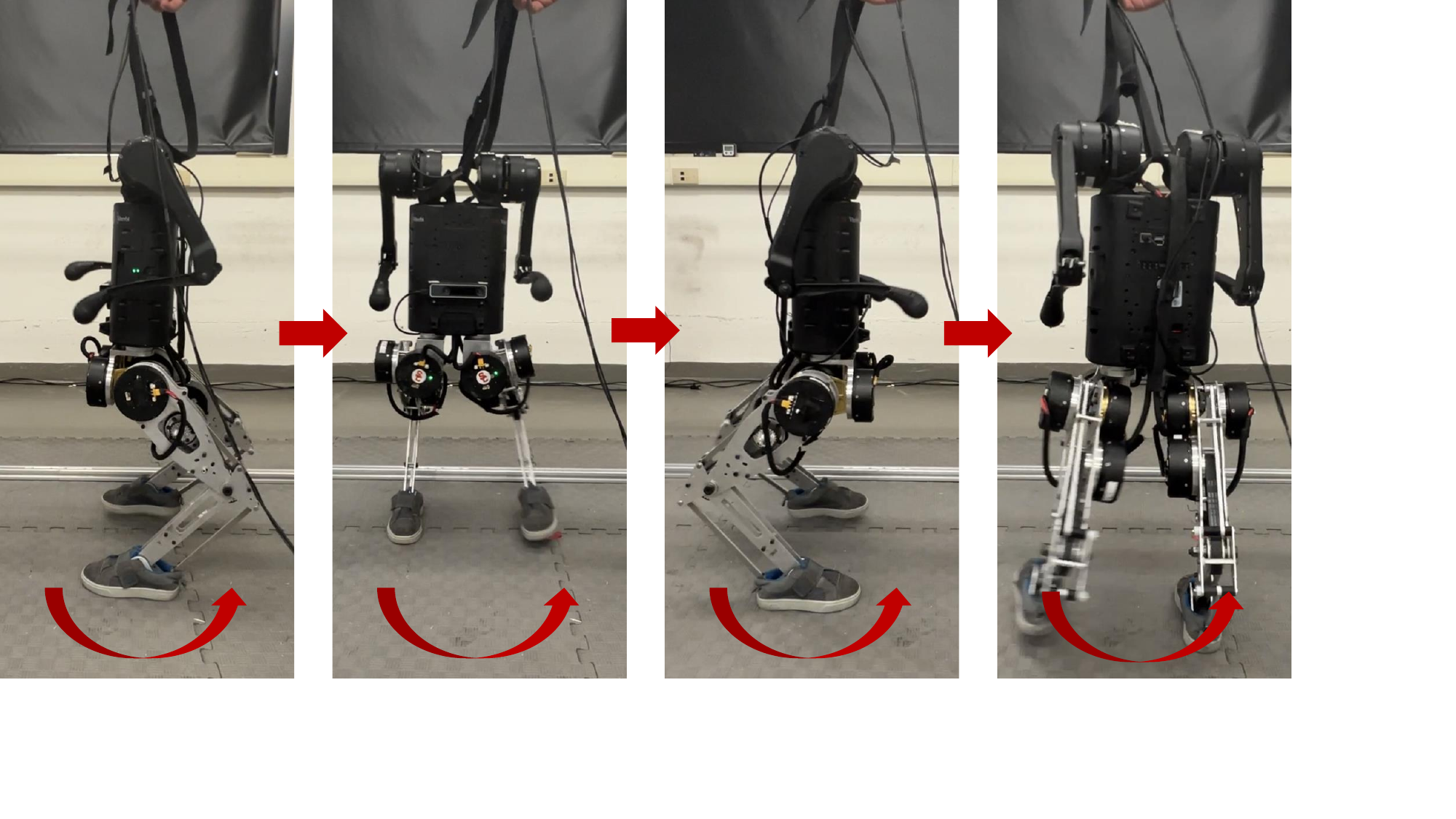}
    \caption{\centering}
    \end{subfigure}
    \\
    \begin{subfigure}[b]{0.49\textwidth}
    \includegraphics[clip, trim=3cm 10.5cm 3cm 10.5cm, width=1\columnwidth]{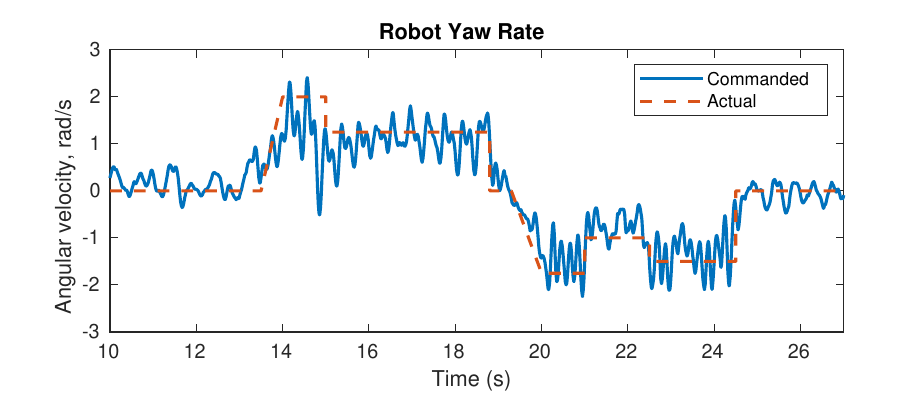}
    \caption{\centering}
    \end{subfigure}
    \caption{Dynamic turning experiment: (a) Turning snapshots in counter-clockwise direction; (b) Yaw rate tracking plot}
    \label{fig:yaw_turning}
\end{figure}

 \subsection{Simulation Results on Dynamic Loco-manipulation}

 Firstly, we present the numerical simulation results on dynamic and contact-informed loco-manipulation control on HECTOR. The custom simulation frameworks of HECTOR are made available as an open-source project\footnote{\url{https://github.com/DRCL-USC/Hector_Simulation}}.  
 In HECTOR's Simulink simulation, our controllers are written in MATLAB scripts. For more efficient computation, we use qpOASES (\cite{ferreau2014qpoases}) in CasADi toolbox (\cite{Andersson2019}) to solve the MPC problems online. 
 In HECTOR's ROS+Gazebo Simulation, the MPC, state estimation, and low-level control scripts are written in C++. MPC problem is solved through qpOASES C++ interface.  

  To demonstrate the effectiveness of the proposed control scheme in loco-manipulation and the importance of contact schedule in MPC, in Simulink simulation, we have allowed the robot to perform multi-contact tasks with heavy objects, shown in Figure \ref{fig:LocoSimSnaps}. The robot can stay well-balanced during stepping even throwing a $2 \: \unit{kg}$ ball. 
  
  In addition, in a logistics setting, we commanded the robot to transfer a $2 \: \unit{kg}$ package from a table to the conveyor belt with in-place walking and turning, illustrated in Figure \ref{fig:90turnComparison}. Compared to the traditional quasi-static approach, where the robot has to perform turning, picking up, and dropping off separately, our proposed dynamic loco-manipulation control can allow the robot to synchronize the locomotion and manipulation tasks altogether, allowing more dynamic and efficient package transferring. 
  
  Furthermore, to further demonstrate the effectiveness of the loco-manipulation SRBD and contact schedule in the proposed MPC, we have leveraged the convenience of the simulation framework for time-varying load. The robot can walk while carrying a time-varying load with mass in the range of $[0.5\:\:4.0 ]\:\unit{kg}$, shown in Figure \ref{fig:loco_snap}. The associated torque plots are shown in Figure \ref{fig:loco_torque}.

\begin{figure}[!t]
     \centering
     \begin{subfigure}[b]{0.5\textwidth}
         \centering
         \includegraphics[clip, trim=0cm 4cm 3.4cm 0cm, width=1\columnwidth]{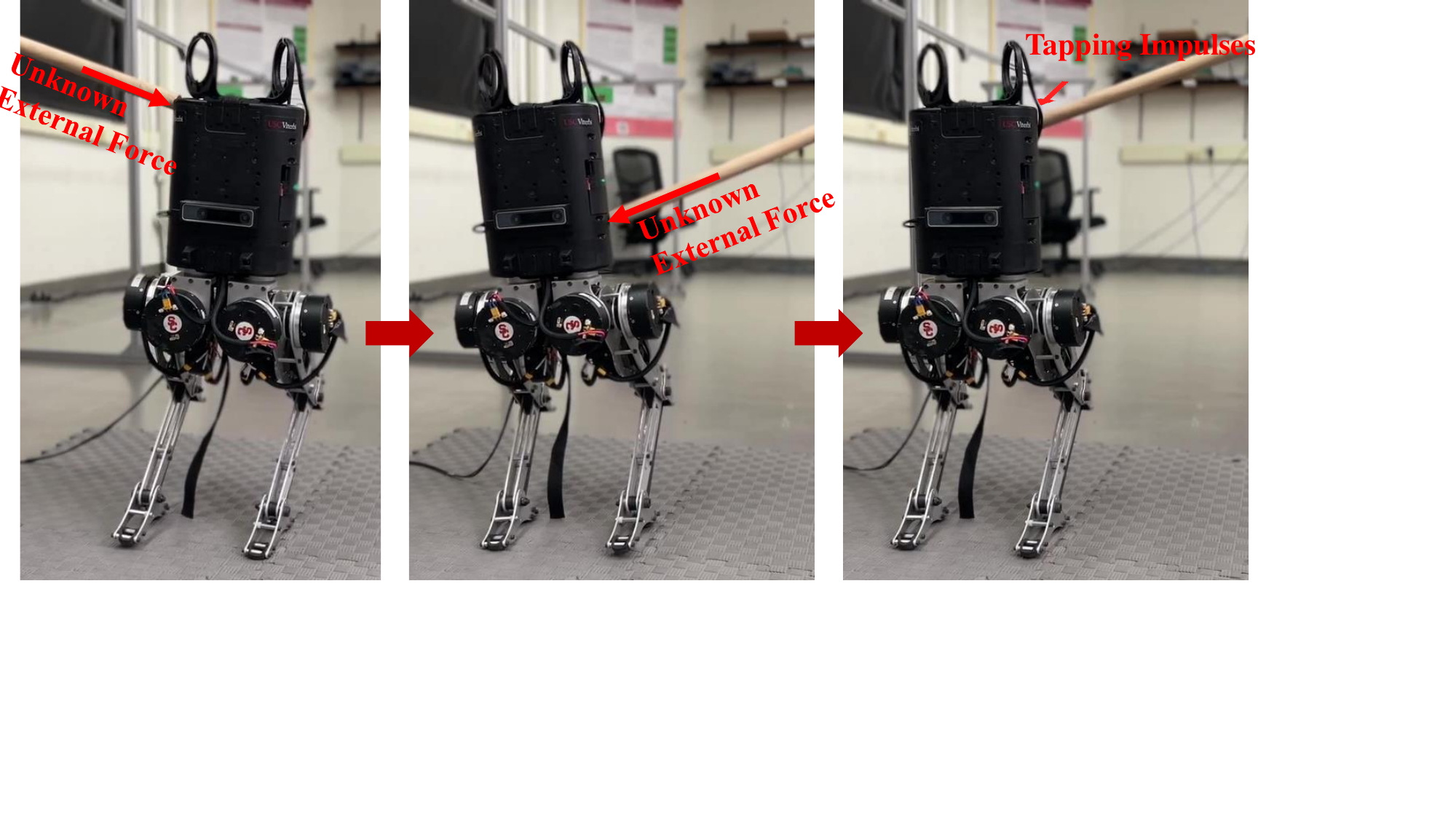}
         \caption{\centering}
         \label{fig:snap_balance}
     \end{subfigure} 
     \vspace{-0.1cm}
     \\
     \begin{subfigure}[b]{0.42\textwidth}
         \centering
         \includegraphics[clip, trim=0cm 0cm 1cm 0cm, width=1\columnwidth]{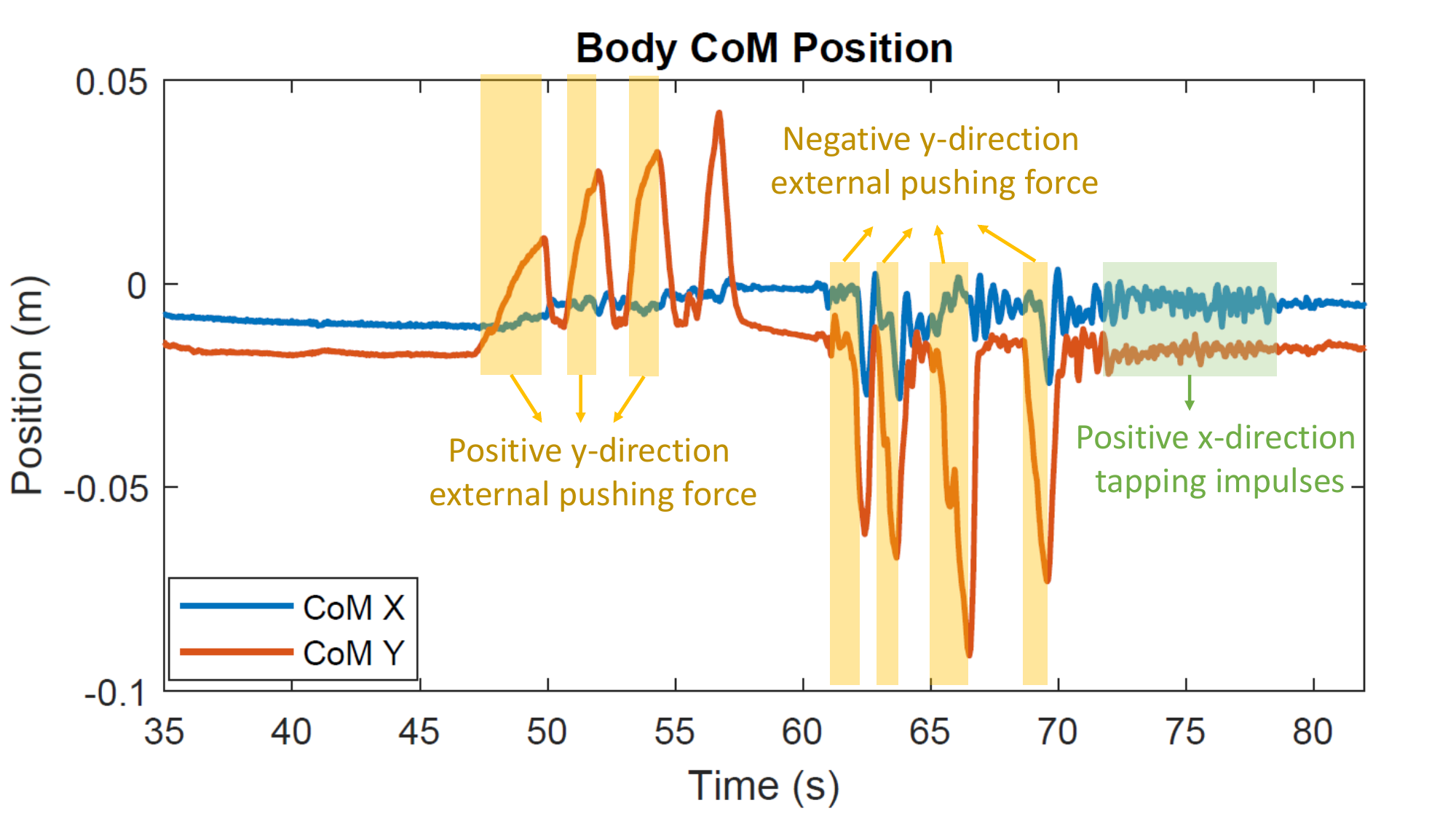}
         \caption{\centering}
         \label{fig:com_balance}
     \end{subfigure}
     \vspace{-0.1cm}
        \caption{ External disturbance rejection in stance: (a) Experiment snapshots; (b) Robot CoM location plots. }
        \label{fig:balance}
        \vspace{-0.2cm}
\end{figure}
\begin{figure}[t]
\vspace{0.2cm}
    \center
    \includegraphics[clip, trim=3cm 9.5cm 3cm 9.5cm, width=0.9\columnwidth]{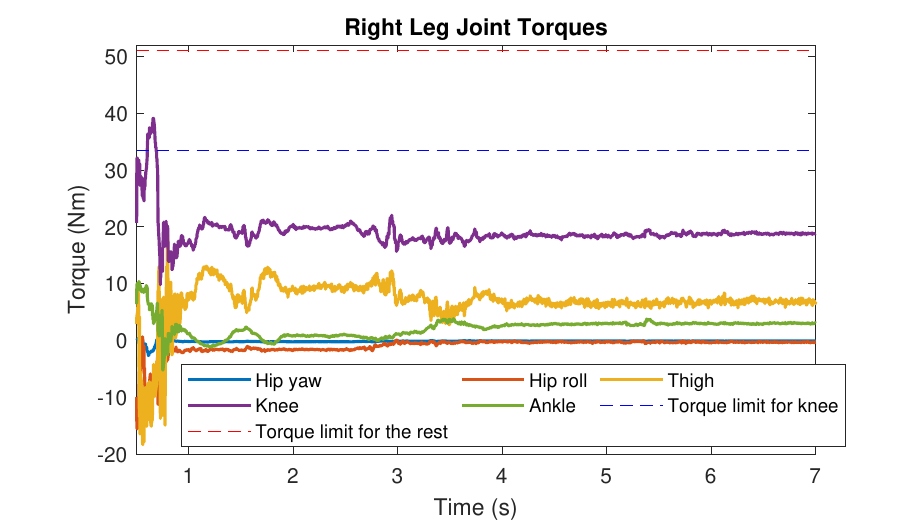}
         \caption{Joint torque plots under 8 kg load in stance.}
         \label{fig:torque_load}
    \vspace{-0.2cm}
\end{figure}

 \subsection{Experimental Results}
 We have also demonstrated the feasibility of our proposed Force-and-moment-based MPC approach in hardware experiments with the HECTOR mini-humanoid robot. 

 \subsubsection{System Integration:}
 Effective state estimation is crucial for achieving accurate and dynamic locomotion in robotics. we address this by integrating a 6-axis IMU in the trunk, positioned near the robot's center of mass (COM) for precise measurements. To estimate the COM states, a linear Kalman filter processes the IMU data and leg kinematics at a high refresh rate of 1 kHz. To enhance the robustness of state estimation, an Intel Realsense T265 tracking camera is also located near the COM. This camera, through its onboard VIO algorithm, outputs the position and orientation of the body at a slower 200 Hz rate. The system ensures accurate and robust state estimation for the HECTOR platform by fusing the data from both sources. Additionally, for potential future applications, HECTOR is equipped with an Intel Realsense D435i camera mounted on top, designed to capture depth data and other relevant environmental information. However, for this work, the robot is unaware of the terrain data.

   \begin{figure}[!t]
\vspace{0.2cm}
     \centering
     \begin{subfigure}[b]{0.5\textwidth}
         \centering
         \includegraphics[clip, trim=4cm 10.2cm 4cm 10.2cm, width=1\columnwidth]{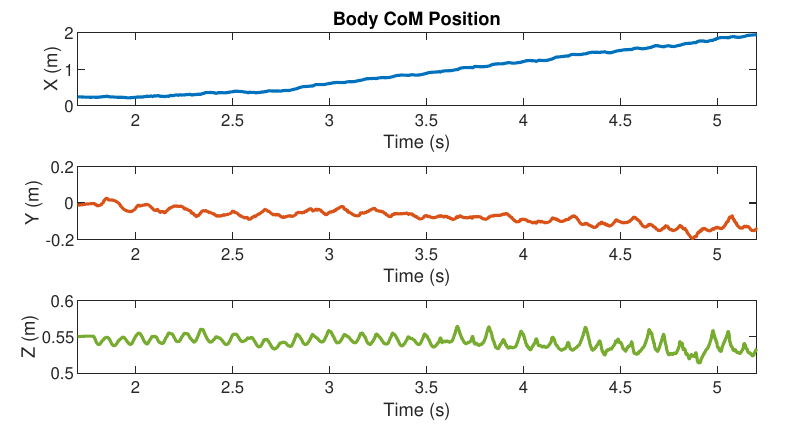}
         \caption{\centering}
         \label{fig:com_hardware1}
     \end{subfigure} 
     \vspace{0.0cm}
     \\
     \begin{subfigure}[b]{0.45\textwidth}
         \centering
         \includegraphics[clip, trim=4cm 9cm 4.5cm 9.3cm, width=1\columnwidth]{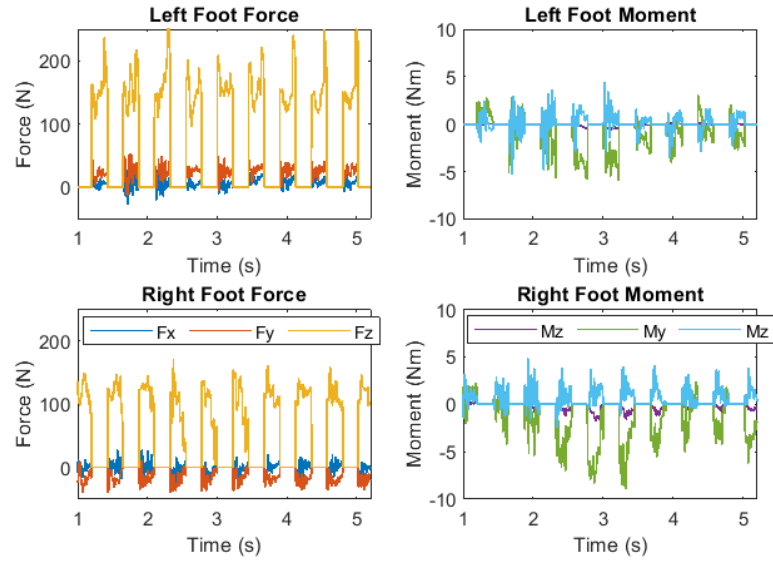}
         \caption{\centering}
         \label{fig:force_hardware1}
     \end{subfigure}
     \vspace{-0.1cm}
        \caption{ Plots of uneven terrain locomotion experiment results with a forward speed of 0.6 m/s: (a) Robot CoM plots; (b) MPC forces and moments solutions. }
        \label{fig:hardware1_plots}
        \vspace{-0.2cm}
\end{figure}

 \subsubsection{Control Tuning:}
 Control objective weight/gain tuning on HECTOR hardware is very minimal to achieve the presented results. The generalized MPC weights used across the tasks are
 \begin{gather}
\nonumber
\bm Q_k = \text{diag}[500\:500\:500\:150\:150\:150\:1\:1\:3\:1\:1\: 1\:0],\: \\
\nonumber
\bm R_k = \text{diag}[1\:1\: 1 \:1\:1\: 1 \:5\:5\: 5 \:5\:5\: 5 \:]\times10^{-3}. \quad \quad \quad \: \: \: 
\end{gather}

\subsubsection{Dynamical Turning:} Firstly, we demonstrate the dynamic turning experiments on HECTOR. Figure \ref{fig:yaw_turning} shows the yaw rate tracking of the robot, in the range of $[-1.75\:\: 2]$ $\unit{rad/s}$. Figure \ref{fig:turning}  showcases the robot's turning capability under the proposed MPC while carrying a 2.5 $\unit{kg}$ payload.

\subsubsection{Handling Heavy Load:}
 To validate the loco-manipulation and load-carrying capabilities of the proposed MPC framework and showcase the power density of HECTOR, we demonstrate the robot's load-carrying ability in both double-leg standing and stepping scenarios. Figure \ref{fig:Carrying} and \ref{fig:Carrying2} shows experiment snapshots of standing while carrying $8 \: \unit{kg}$ load ($50\%$ robot mass), and stepping while carrying $4 \: \unit{kg}$ load. The joint torques associated with the experiments carrying $8 \: \unit{kg}$ during stance are shown in Figure \ref{fig:torque_load}. Note that the joint torques in the load-carrying experiment are well under the peak torque limit of the robot joints. 

 \begin{figure}[!t]
\vspace{0.2cm}
     \centering
     \begin{subfigure}[b]{0.45\textwidth}
         \centering
         \includegraphics[clip, trim=4cm 8.5cm 4.5cm 8.5cm, width=1\columnwidth]{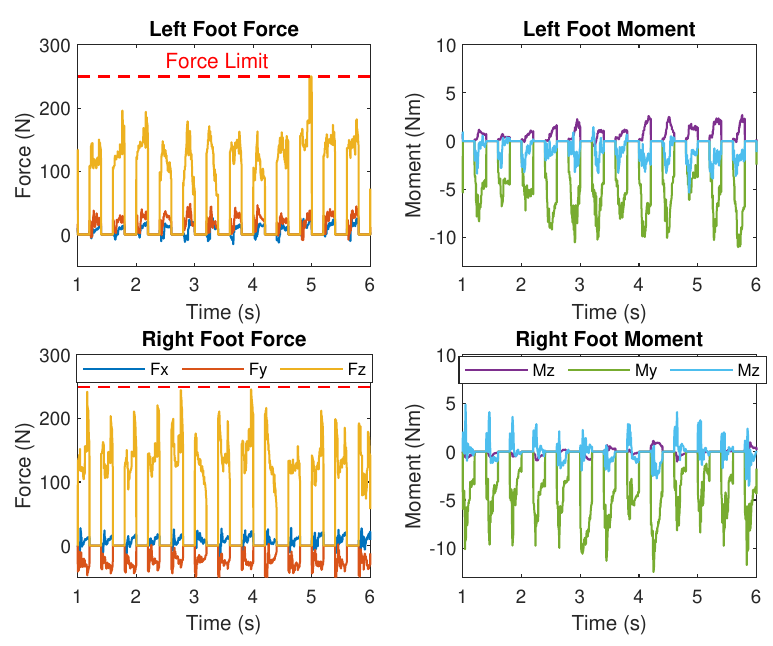}
         \caption{MPC Forces and Moments}
         \label{fig:force_hardware2}
     \end{subfigure}
    \\
     \begin{subfigure}[b]{0.45\textwidth}
         \centering
         \includegraphics[clip, trim=4cm 10cm 4.5cm 10cm, width=1\columnwidth]{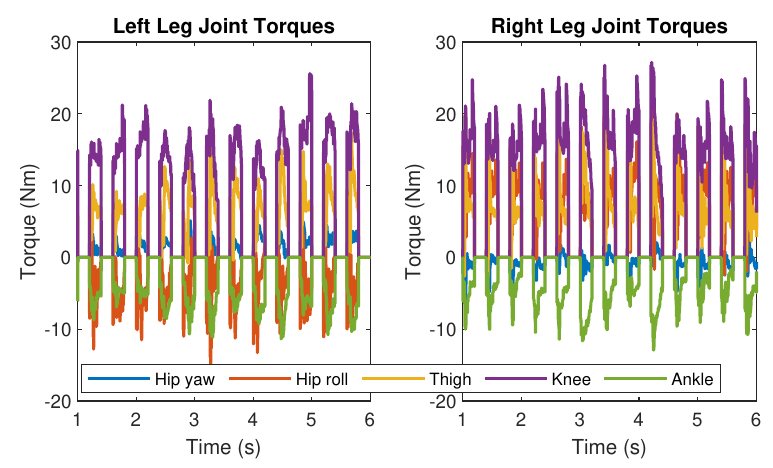}
         \caption{Stance Joint Torques by MPC Forces and Moments}
         \label{fig:torque_hardware2}
     \end{subfigure}
     \vspace{-0.1cm}
        \caption{ Plots of uneven terrain loco-manipulation experiment results with 2.5 kg payload: (a) MPC forces and moments solutions. (b) Stance leg joint torques by MPC forces and moments solutions. }
        \label{fig:hardware2_plots}
        \vspace{-0.4cm}
\end{figure}

 \subsubsection{Disturbance Rejection in Balancing:}
 
 The proposed Force-and-moment-based MPC approach also demonstrates robustness in recovering from external perturbations including unknown external forces and unknown terrain changes during standing configuration. Figure \ref{fig:balancing} shows the snapshots of HECTOR's robust balancing capability in stance while constantly moving its feet on a seesaw. Figure \ref{fig:balance} shows the experimental results of balancing while applying unknown forces and impulses to HECTOR. It can be observed that HECTOR can recover quickly to the desired CoM position after external forces and impulses are applied.

 \subsubsection{Robust Multi-terrain Locomotion:}

The locomotion ability of HECTOR robot with Force-and-moment-based MPC is also promising. We have demonstrated uneven terrain locomotion on HECTOR hardware in both biped and humanoid forms. Figure \ref{fig:expSnaps} presents experiment snapshots of uneven terrain locomotion with different terrain setups, including 
\begin{itemize}
    \item Stacked wood slats up to $6\: \unit{cm}$, shown in Figure \ref{fig:biped_terrain1} and \ref{fig:humanoid_terrain1};
    \item Randomly placed wood slats, shown in Figure \ref{fig:biped_terrain2};
    \item Randomly placed wood slats on wet grassy terrain, shown in Figure \ref{fig:biped_grass};
    \item 18$^\circ$ slope, shown in Figure \ref{fig:slope}.
\end{itemize}
Our MPC control scheme on HECTOR can confidently traverse the above terrain setups. Figure \ref{fig:hardware1_plots} shows the CoM location and MPC forces and moments (in world frame) plots of the HECTOR walking over stacked wood slats with $0.6\: \unit{m/s}$ forward speed (illustrated in snapshots in Figure \ref{fig:biped_terrain1}). The wood slats have heights of $2\: \unit{cm}$, $4\: \unit{cm}$, and $6\: \unit{cm}$. 

\subsubsection{Dynamic Loco-manipulation}

\rev{Furthermore, we have achieved dynamic loco-manipulation on HECTOR humanoid robot. Figure \ref{fig:turning} shows the snapshots of HECTOR turning dynamically while carrying 2.5 $\unit{kg}$ payload. Figure \ref{fig:loco_hardware} shows the snapshots of HECTOR walking over randomly placed and unstable wood slats with a 2.5 $\unit{kg}$ payload (composed of 2 Unitree Go1 batteries). The dynamic loco-manipulation MPC solutions (in world frame) and associated joint torques are shown in Figure \ref{fig:hardware2_plots}. Note that all force solutions comply with the force limit set in MPC  (250 $\unit{N}$) and the associated joint torques are well under the joint torque limits of the hardware. }

\section{Future Work}
\label{sec:futureWork}

Having outlined the primary experimental validations for the proposed control schemes, we aim to highlight future research directions to enhance autonomous loco-manipulation. 

Currently, the controller requires manual inputs of the CoM and mass properties of the carried object. However, we are exploring a more autonomous approach involving vision-based object shape detection and object state estimation to determine the object's CoM position. 

Additionally, we intend to estimate and sense the object's weight autonomously, which is particularly important in the problem of carrying an unknown and time-varying load. The first future integration is by using the upper-limb motors. The back-drivability feature of these QDD motors plays a vital role here, allowing us to sense the object's weight through the back-drive current/torque of these motors and the kinematic pose of the upper limb. The second future research direction to tackle this problem revolves around using L$_1$ adaptive control with MPC for compensating the unknown load. Successful integration and implementation have been seen on quadrupedal robots, such as in \cite{sombolestan2023adaptive}. 


\section{Conclusions}
\label{sec:Conclusion}




\rev{In summary, this work presents an effective approach to dynamic locomotion and loco-manipulation control on humanoid robots through a generalized Force-and-Moment-Based Model Predictive Control (MPC). Our proposed framework involves a simplified rigid body dynamics (SRBD) model, accounting for both humanoid and object dynamics in loco-manipulation. 
we have demonstrated the effectiveness of the proposed control framework on  HECTOR—a dynamic, small-scale, power-dense, and cost-effective humanoid robot platform. HECTOR demonstrates exceptional balance in double-leg stance, even under external force disturbances, and executes dynamic walking across varied terrains at speeds up to 0.6 m/s. Furthermore, we achieve dynamic loco-manipulation over uneven terrain, carrying a 2.5 kg load. HECTOR simulation framework and controllers are made available as an open-source project on GitHub.}

\section{Acknowledgment}
\label{sec:Acknowledgement}

The authors would like to thank Yiyu Chen, Zhanhao Le, and Han Gong for contributing to the HECTOR open-source software. 

\section{Funding}
\label{sec:funding}
This work is supported by the USC departmental startup fund. 

\bibliographystyle{sageH}
\bibliography{reference.bib}

\end{document}